\pdfoutput=1
\documentclass[conference]{IEEEtran}

\usepackage{times}
\usepackage{soul}
\usepackage{multirow}
\usepackage{multicol}
\usepackage{microtype, breqn}
\usepackage{titlesec}
\usepackage[dvipsnames]{xcolor}
\usepackage{afterpage}
\usepackage{physics}
\usepackage{comment}
\usepackage{amsfonts}
\usepackage{graphicx}
\usepackage{subfloat}
\usepackage{amsmath}
\usepackage{subcaption}
\usepackage{dsfont}
\usepackage{fancyhdr}
\usepackage{cite}
\usepackage{amssymb}

\usepackage{amsthm}
\usepackage[bookmarks=true]{hyperref}

\usepackage[utf8]{inputenc}
\usepackage{bm}
\usepackage{upgreek} 
\let\theta\uptheta 

\def\QED{\hfill \quad{\bf Q.E.D.}\medskip}

\newcounter{yaocounter}

\newcommand{\yao}[1]{\textcolor{blue}{#1}}

\newtheorem{theorem}{Theorem}

\newtheorem{corollary}{Corollary}

\begin{document}

\title{Enabling Team of Teams: \\A Trust Inference and Propagation (TIP) Model in \\Multi-Human Multi-Robot Teams}

\author{
\authorblockN{Yaohui Guo, X. Jessie Yang, and Cong Shi}
\authorblockA{
Industrial and Operations Engineering\\
University of Michigan\\
Ann Arbor, Michigan, USA\\
\texttt{\{yaohuig, xijyang, shicong\}@umich.edu}
}}

\maketitle


\begin{abstract}
Trust has been identified as a central factor for effective human-robot teaming. Existing literature on trust modeling predominantly focuses on dyadic human-autonomy teams where one human agent interacts with one robot. There is little, if not no, research on trust modeling in teams consisting of multiple human agents and multiple robotic agents.
To fill this research gap, we present the trust inference and propagation (TIP) model for trust modeling in multi-human multi-robot teams. In a multi-human multi-robot team, we postulate that there exist two types of experiences that a human agent has with a robot: direct and indirect experiences. The TIP model presents a novel mathematical framework that explicitly accounts for both types of experiences. To evaluate the model, we conducted a human-subject experiment with 15 pairs of participants (${N=30}$). Each pair performed a search and detection task with two drones.  Results show that our TIP model successfully captured the underlying trust dynamics and significantly outperformed a baseline model. To the best of our knowledge, the TIP model is the first mathematical framework for computational trust modeling in multi-human multi-robot teams.

\end{abstract}


\IEEEpeerreviewmaketitle

\section{Introduction}\label{sec:intro}

A human agent's trust in an autonomous/robotic agent, defined as ``the attitude that an agent will help achieve an individual’s goals in situations characterized by uncertainty and vulnerability~\cite{See:2004vj}'', is a central factor for effective human-robot interaction (HRI)~\cite{sheridan_humanrobot_2016, Yang2021_HFJ, Soh-RSS-18}. 
Optimal interaction can be achieved only when an appropriate level of trust is established between the human and the robotic agents~\cite{parasuraman1997humans, hancock2021evolving}. 
Despite the extensive research efforts over the past thirty years, existing literature is predominantly focused on trust modeling in dyadic human-robot teams where one human agent interacts with one robot~\cite{NAP26355}. There is little, if not no, research on trust modeling in teams consisting of multiple human agents and multiple robots.

\begin{figure}[h]
  \begin{center}
  \vspace{0pt}
    \includegraphics[width=1\columnwidth]{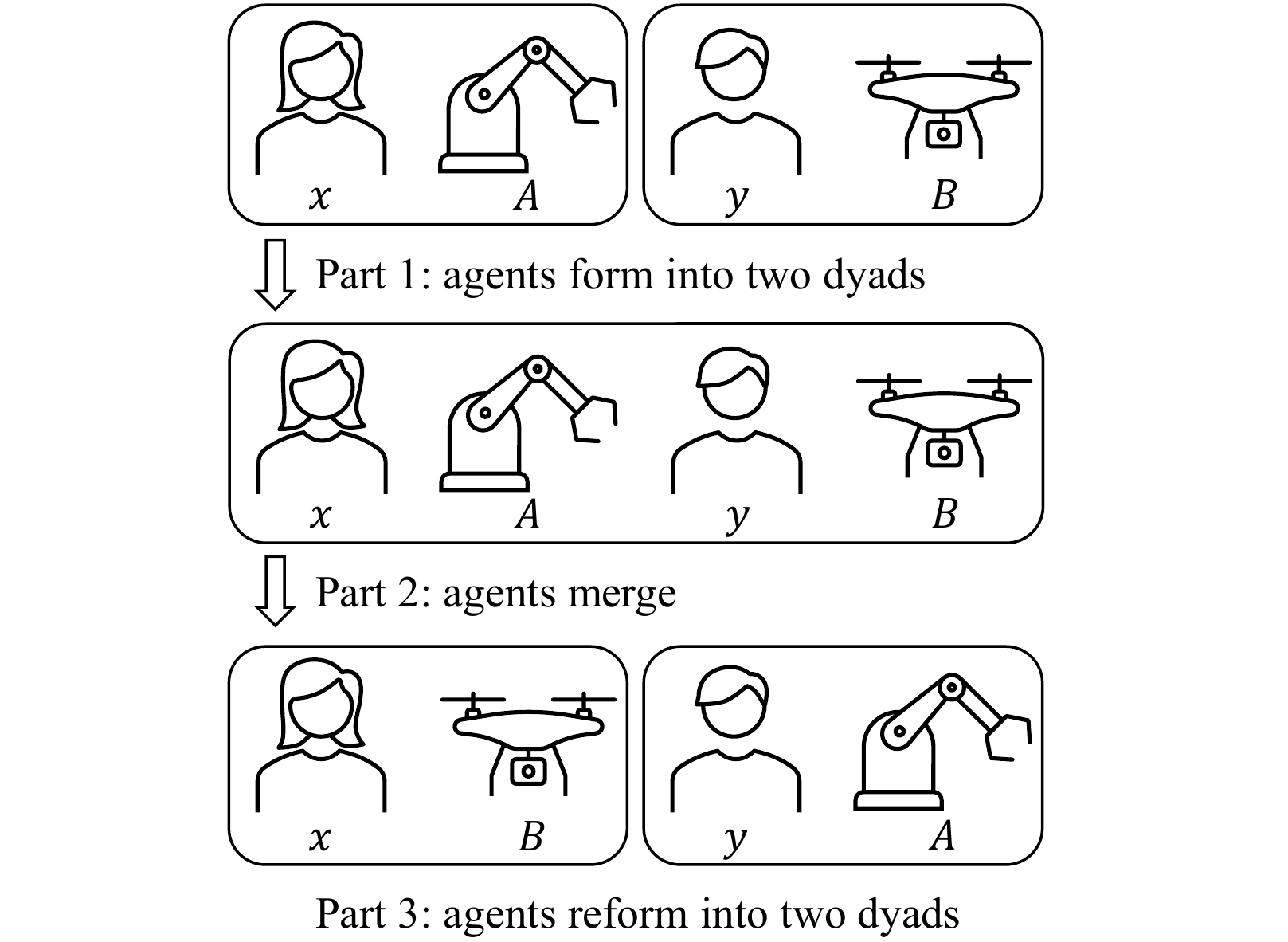}
  \end{center}
  \caption{Four agents can form sub-teams. In part 1, human $x$ and robot $A$ form a dyad, and human $y$ and robot $B$ form a dyad. In part 2, two dyads merge. In part 3, human $x$ and robot $B$ form a dyad, and human $y$ and robot $A$ form a dyad.}
  \label{fig:scenario}
\end{figure}

Consider a scenario where two human agents (figure \ref{fig:scenario}), $x$ and $y$, and two robots, $A$ and $B$, are to perform a task. The four agents are allowed to form sub-teams to enhance task performance (e.g., maximizing throughput and minimizing task completion time). For instance, they could initially form two dyadic human-robot teams to complete the first part of the task, merge to complete the second part and split again with a different configuration to complete the third part of the task, and so on. This scenario illustrates a new organizational model known as ``team of teams~\cite{Meehan2018, McChrystal}'' in which the team composition is fluid and team members come and go as the nature of the problem changes.

\begin{figure}[t]
  \centering
  \includegraphics[width=1\linewidth]{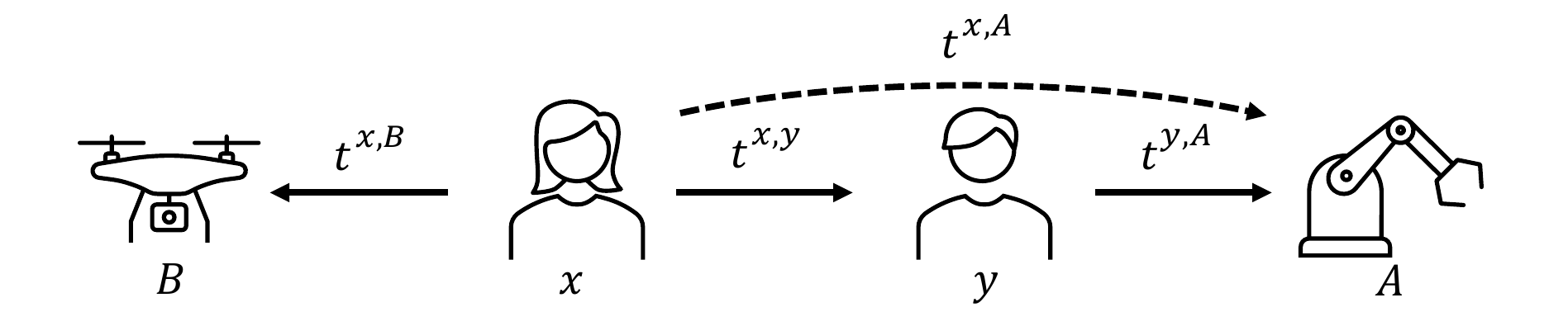}
  \vspace{-12pt}
  \caption{An arrow points from a trustor to a trustee, representing the trust $t^{\text{trustor}, \text{trustee}}$. Human $x$ updates her trust in robot $B$ based on direct experience. 
  Even though $x$ does not have direct interaction with $A$, $x$ could still update her trust toward $A$ through a third party, $y$. }
  \label{fig:trust-transfer}
\end{figure}

In this scenario, we postulate that there exist two types of experiences that a human agent has with a robot: \emph{direct and indirect experiences}. Direct experience, by its name, means that a human agent's interaction with a robot is by him-/her-self; indirect experience means that a human agent's interaction with a robot is mediated by another party. Considering the third part of the task (see figures \ref{fig:scenario} and \ref{fig:trust-transfer}), human $x$ works directly with robot $B$ (i.e., direct experience). Even though there is no direct interaction between $x$ and $A$ in part 3, we postulate that $x$ could still update his or her trust in $A$ by learning her human teammate $y$'s experience with $A$, i.e., $y$'s direct experience with $A$ becomes $x$'s \textit{in}direct experience with $A$, based on which $x$ can update her trust in $A$, $t^{x, A}$. Essentially, $y$'s trust in A \textit{propogates} to $x$.


Under the direct and indirect experience framework, prior work on trust modeling in dyadic human-robot teams can be regarded as examining how \emph{direct} experience influences a person's trust in a robot. In multi-human multi-robot teams, we postulate that \textit{both direct and indirect experiences drive a human agent's trust in a robot}.

In this study, we develop the \textbf{T}rust \textbf{I}nference and \textbf{P}ropagation (TIP) model for multi-human multi-robot teams. 
The proposed model explicitly accounts for the direct and indirect experiences a human agent may have with a robot. We examine trust dynamics under the TIP framework and prove theoretically that trust converges after repeated (direct and indirect) interactions. To evaluate the proposed TIP model, we conducted a human-subject experiment with 15 pairs of participants ($N=30$). Each pair worked with two drones to perform a threat detection task for 15 sessions. We compared the TIP model (i.e., accounts for both the direct and indirect experiences) and a direct-experience-only model (i.e., only accounts for the direct experience a human agent has with a robot). Results show that the TIP model successfully captures people's trust dynamics with a significantly smaller root-mean-square error (RMSE) compared to the direct-experience-only model. 



\smallskip
The key contribution of this work is three-fold:
\begin{itemize}
    \item To the best of our knowledge, the proposed TIP model is the first mathematical framework  for computational trust modeling in multi-human multi-robot teams. The TIP model accounts for both the direct and indirect experiences (through trust propagation) a human agent has with a robot in multi-human multi-robot teams. As a result, the TIP model is well-suited for trust estimation in networks involving multiple humans and robots.
    \item We prove theoretically that trust converges to the unique equilibrium in probability after repeated direct and indirect interactions under our TIP framework. Such an equilibrium can also be efficiently computed.
    \item We conduct a human-subject experiment to assess the TIP model. Results reveal the superior performance of the TIP model in capturing trust dynamics in a multi-human multi-robot team.  

\end{itemize}

This paper is organized as follows. Section~\ref{sec:relatedWork} presents related work, including trust modeling in dyadic human-robot teams and reputation/trust management in e-commerce. In Section~\ref{sec:model}, we describe the mathematical formulation of the TIP model and examine its behavior under different types of interactions. Section~\ref{sec:humanStudy} presents the human-subject study. In Section~\ref{sec:results_discuccion}, we present and discuss the results. Section~\ref{sec:conclusion} concludes the paper.

\section{Related Work}\label{sec:relatedWork}
In this section, we review two bodies of research motivating the present study: the extensive literature on trust in dyadic human-robot teams and the literature on reputation/trust management. The latter is a research topic in computer science that shares commonalities with the underlying research question of trust modeling in multi-human multi-robot teams.



\subsection{Trust Modeling in Dyadic Human-robot Interaction}\label{sec:HRI_trust_review}

Trust in autonomous/robotic agents attracts research attention from multiple disciplines. One line of research is to identify factors influencing a human's trust in autonomy/robots and quantify their effects. These factors can be categorized into human-related factors such as personality~\cite{Bhat_RAL_2022}, robot-related factors such as reliability~\cite{Lyons_HFJ_2021, gombolay_robotic_2018} and transparency~\cite{Wang:2016ui, luo_evaluating_2022}, and task-related factors such as task emergency~\cite{Robinette:2016_HRI}. For a review of the factors, see~\cite{hancock2021evolving}. 
More recently, another line of research has emerged that focuses on understanding the dynamics of trust formation and evolution when a person interacts with autonomy repeatedly~\cite{Yang2021_HFJ, DeVisser_IJSR, Guo2020_IJSR}. Empirical studies have investigated how trust strengthens or decays due to moment-to-moment interactions with autonomy~\cite{Lee:1992it, manzey2012human, Yang2021_HFJ, Yang:2017:EEU:2909824.3020230}. Based on the empirical research, three major properties of trust dynamics have been identified and summarized, namely \textit{continuity}, \textit{negativity bias}, and \textit{stabilization} \cite{Guo2020_IJSR, Yang2021_chapter} 

Acknowledging that trust is a dynamic variable, several computational trust models in dyadic human-robot teams have been developed~\cite{chen2018planning,  Xu2015optimo,  Guo2020_IJSR, wang2016impact}. Notably, Xu and Dudek~\cite{Xu2015optimo} proposed the online probabilistic trust inference model (OPTIMo) utilizing Bayesian networks to estimate human trust based on the autonomous agent's performance and human behavioral signals.
Guo and Yang~\cite{Guo2020_IJSR} modeled trust as a Beta random variable parameterized by positive and negative interaction experience a human agent has with a robotic agent. Soh et al.~\cite{soh2020multi} proposed a Bayesian model which combines Gaussian processes and recurrent neural networks to predict trust over different tasks. For a detailed review, refer to~\cite{kok2020trust}.

\subsection{Reputation/Credential Management} 


Despite limited research on trust modeling in multi-human multi-robot teams, insights can be drawn from studies on reputation management. In consumer-to-consumer electronic marketplaces like eBay, reputation systems play a crucial role in generating trust among buyers to facilitate transactions with unknown sellers~\cite{Dellarocas2003}. These systems can be categorized as centralized, where reputation values are stored centrally, representing the overall trustworthiness of sellers, or decentralized, where buyers maintain their evaluation scores privately~\cite{hendrikx2015reputation}. In decentralized systems, a propagation mechanism allows buyers to obtain reputation values, even in the absence of prior transactions. Various propagation mechanisms have been developed, such as subjective logic integrated into the Beta reputation management system~\cite{josang1997artificial, josang2002beta} or the concept of "witness reputation" in the FIRE reputation management model~\cite{huynh2004fire}, facilitating the transfer of reputation scores among agents in a network. These propagation mechanisms provide valuable insights into modeling trust update through indirect experience in HRI. Yet, their direct application to HRI settings is impeded due to the distinct characteristics of human trust in robots, as reviewed in Section~\ref{sec:HRI_trust_review}.

\section{Mathematical Model}\label{sec:model}
We present the TIP model in this section. Our key motivation is to develop a fully computational trust inference and propagation model that works in general multi-human multi-robot settings. First, we discuss the assumptions and introduce the mathematical formulation. Second, we examine the behavior of the model under repeated human-robot interactions. Finally, we present the parameter inference method and trust estimation using the TIP model. 



\subsection{Assumptions}\label{subsec:assumptions}

We make three major assumptions in the context of HRI. First, we assume that \textit{each human agent communicates trust as a single-dimensional value}. In some prior work, trust is represented as a tuple. For example, in~\cite{josang2002beta}, trust is represented as a triplet, i.e., belief, disbelief, and uncertainty. Although a multi-dimensional representation conveys more information, our study as well as some prior studies show that a one-dimensional representation of trust suffices in capturing trust evolution ~\cite{chen2018planning, Xu2015optimo, wang2016impact, Guo2020_IJSR, guo2021reverse}.  
Moreover, querying a single-dimension trust value increases operational feasibility because keeping track of multiple numbers adds unnecessary cognitive load and may not be pragmatic for non-experts. Therefore, we assume a simple one-dimension form of trust in this study.


Second, we assume that \textit{the human agents are cooperative}, i.e., they are honest and willing to share their trust in a robot truthfully with their human teammates. 

Third, we take an ability/performance-centric view of trust and assume that a human agent's trust in a robot is primarily driven by the ability or performance of the robot. This ability/performance-centric view has been widely used in prior research for modeling trust in task-oriented HRI contexts (i.e., a robot is to perform a specific task)~\cite{Xu2015optimo, chen2018planning, Guo2020_IJSR}. 

We discuss the limitations of the assumptions in Section \ref{sec:conclusion}.





\subsection{Proposed Model}

\noindent \textbf{Trust as a Beta random variable.} We take a probabilistic view to model trust as in~\cite{Guo2020_IJSR}. At time $k$, the trust $t_{k}^{a,b}$ that a human agent $a$ feels toward another agent $b$ follows a Beta distribution, i.e., 
\begin{equation} \label{eq:trust_def} 
t_{k}^{a,b} \sim \operatorname{Beta}\left( \alpha _{k}^{a,b} ,\beta _{k}^{a,b}\right)
\text{,}
\end{equation}
where $\alpha _{k}^{a,b}$ and $\beta _{k}^{a,b}$ are the positive and negative experiences $a$ had about $b$ up to time $k$, respectively, $k=0,1,2,\dots$. When $k=0$, $\alpha _{0}^{a,b}$ and $\beta _{0}^{a,b}$ represent the prior experiences that $a$ has before any interaction with $b$. The expected trust is given by
\begin{equation} \label{eq:trust_expected} 
\mu _{k}^{a,b} =\alpha _{k}^{a,b} /\left( \alpha _{k}^{a,b} +\beta _{k}^{a,b}\right)
\text{.}
\end{equation}
Here we note that $t_{k}^{a,b}$ is the queried trust given by the agent $a$, which has some randomness due to subjectivity, while $\mu_{k}^{a,b}$ is the expected trust determined by the experiences.


\smallskip
\noindent \textbf{Trust update through direct experience.} Similar to~\cite{Guo2020_IJSR}, we update the experiences through direct interaction at time ${k}$ by setting 
\begin{equation}\label{eq:direct_update}
\begin{aligned}
\alpha _{k}^{a,b} & =\alpha _{k-1}^{a,b} +s^{a,b} \cdot p_{k}^{b}\\
\beta _{k}^{a,b} & =\beta _{k-1}^{a,b} +f^{a,b} \cdot \overline{p}_{k}^{b}
\end{aligned}\text{.}
\end{equation}
Here $p_{k}^{b}$ and $\overline{p}_{k}^{b}$ are the measurements of $b$'s success and failure during time $k$, respectively; $s^{a,b}$ and $f^{a,b}$ are $a$'s unit experience gains with respect to success or failure of $b$. We require $s^{a,b}$ and $f^{a,b}$ to be positive to ensure that cumulative experiences are non-decreasing. The updated trust $t_{k}^{a,b}$ follows the distribution $\operatorname{Beta} (\alpha _{k}^{a,b} ,\beta _{k}^{a,b})$.

\smallskip
\noindent \textbf{Trust update through indirect experience propagation.} Let $x$ and $y$ denote two human agents and let $A$ denote a robot agent, as illustrated in figure~\ref{fig:trust-transfer}. At time $k$, $y$ communicates his or her trust $t_{k}^{y,A}$ in $A$ with $x$, and then $x$ updates his or her experiences through indirect interaction by 
\begin{equation}\label{eq:indirect_update}
\begin{aligned}
\alpha _{k}^{x,A} & =\alpha _{k-1}^{x,A} +\hat{s}^{x,A} \cdot t_{k}^{x,y} \cdot \left[ t_{k}^{y,A} -t_{k-1}^{x,A}\right]^{+}\\
\beta _{k}^{x,A} & =\beta _{k-1}^{x,A} +\hat{f}^{x,A} \cdot t_{k}^{x,y} \cdot \left[ t_{k-1}^{x,A} -t_{k}^{y,A}\right]^{+}
\end{aligned}\text{,}
\end{equation}
where the superscript `$+$' means taking the positive part of the corresponding number, i.e., $t^+=\max\{0,t\}$ for a real number $t$, and $t_{k}^{x,A}\sim \operatorname{Beta} ( \alpha _{k}^{x,A} ,\beta _{k}^{x,A} )$. 

The intuition behind this model is that $x$ needs to reason upon $t_{k}^{y,A}$, i.e., $y$'s trust toward $A$. First, $x$ compares $y$'s trust $t_{k}^{y,A}$ with his or her previous trust $t_{k-1}^{x,A}$. Let $\Delta t:=t_{k}^{y,A} -t_{k-1}^{x,A}$ be the difference. If $\Delta t \ge 0$, $x$ gains positive indirect experience about $A$, which amounts to the product of the trust difference $\Delta t$, a coefficient $\hat{s}^{x,A}$, and a discounting factor $t_{k}^{x,y}$, i.e., $x$'s trust in $y$; if $\Delta t<0$, then $x$ gains negative indirect experience about $A$, which is defined similarly. 

\subsection{Asymptotic Behavior under Repeated Interactions}
\begin{figure}[h]
  \centering
  \includegraphics[width=1\linewidth]{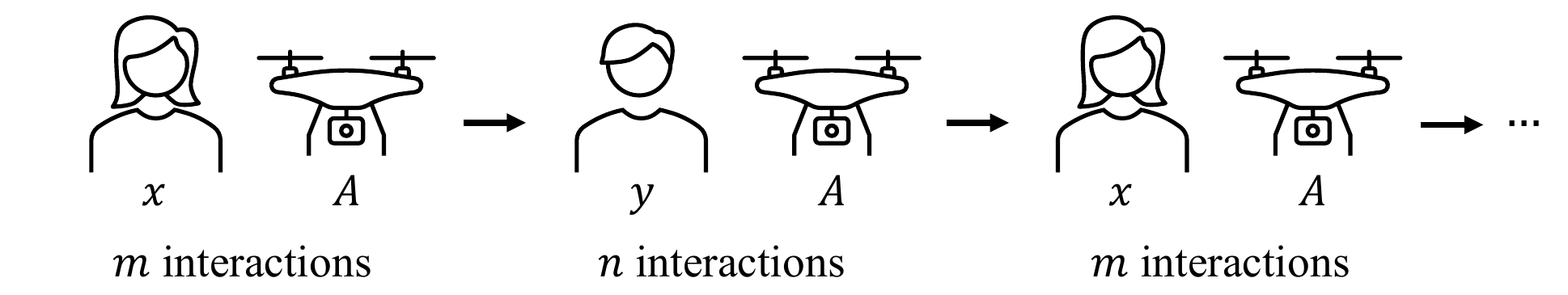}
  \caption{$x$ and $y$ take turns to interact with $A$.}
  \label{fig:repetitive_interactions}
\end{figure}

We examine the behavior of the proposed model under both direct and indirect trust updates. Consider a scenario where human agents $x$ and $y$ take turns working with robot $A$ repetitively. Suppose each $x$'s turn contains $m$ interactions while each $y$'s turn contains $n$ interactions; and, after each interaction, the agent who works directly with $A$ informs the other agent of his or her trust in $A$. Figure~\ref{fig:repetitive_interactions} illustrates the interaction process. In addition, we assume that robot $A$ has constant reliability $r$, i.e., $A$'s performance measure are $p_{k}^A =r$ and $\overline{p}^{A}_k = \bar{r}$, for $k=1,2,\dots,K$, where $\bar{r}:=1-r$, and $x$ has constant trust $t^{x,y}$ in $y$. To avoid triviality, we exclude the case when $m=n=0$ (where no interactions occur). Without loss of generality, we assume $m>0$ and $n\geqslant0$. (The case $m\geqslant0$ and $n>0$ is symmetric.)

We have the following main result on the asymptotic behavior of $t_{k}^{x,A}$ and $t_{k}^{y,A}$.
\begin{theorem}
\label{thm:both_converge}
When $m >0$ and $n\geqslant 0$, $t_{k}^{x,A}$ and $t_{k}^{y,A}$ converge in probability (i.p.) respectively, i.e., there exists $t^{x}$ and $t^{y}$ such that, for any $\epsilon  >0$,
\begin{equation*}
\begin{aligned}
 &  & \lim _{k\rightarrow \infty } \Pr\left(\left| t_{k}^{x,A} -t^{x}\right|  >\epsilon \right) =0\ \\
\text{and} &  & \lim _{k\rightarrow \infty } \Pr\left(\left| t_{k}^{y,A} -t^{y}\right|  >\epsilon \right) =0.
\end{aligned}
\end{equation*}
\end{theorem}
Theorem \ref{thm:both_converge} exhibits that, under alternating interactions with the robot, both agents' trust will stabilize and converge after sufficiently many interactions. The next result gives an exact method to compute the limiting equilibrium.

\begin{theorem}
\label{thm:equilibria}  
The equilibrium $t^{x}$ and $t^{y}$ in Theorem~\ref{thm:both_converge}  satisfy
\begin{equation}
\label{eq:equilibria_case1}
\begin{aligned}
S^{x}\frac{1-t^{x}}{t^{x}} & =\hat{F}^{x}\left( t^{x} -t^{y}\right) +F^{x} & \text{and}\\
F^{y}\frac{t^{y}}{1-t^{y}} & =\hat{S}^{y}\left( t^{x} -t^{y}\right) +S^{y} , & 
\end{aligned}
\end{equation}
if $S^{x} F^{y} \geqslant F^{x} S^{y}$; otherwise, they satisfy
\begin{equation}
\label{eq:equilibria_case2}
\begin{aligned}
F^{x}\frac{t^{x}}{1-t^{x}} & =\hat{S}^{x}\left( t^{y} -t^{x}\right) +S^{x} & \text{and}\\
S^{y}\frac{1-t^{y}}{t^{y}} & =\hat{F}^{y}\left( t^{y} -t^{x}\right) +F^{y} , & 
\end{aligned}
\end{equation}
where $\hat{S}^{x} =nt^{x,y}\hat{s}^{x,A}$, $\hat{F}^{x} =nt^{x,y}\hat{f}^{x,A}$, $S^{x} =ms^{x,A} r$, $F^{x} =mf^{x,A}\overline{r}$, $\hat{S}^{y} =mt^{y,x}\hat{s}^{y,A}$, $\hat{F}^{y} =mt^{y,x}\hat{f}^{y,A}$, $S^{y} =ns^{y,A} r$, and $F^{y} =nf^{y,A}\overline{r}$.
\end{theorem}
The capitalized variables in Theorem~\ref{thm:equilibria} are related to the average experience gains in the long run, e.g., ${S}^{x}$ is $x$'s direct positive experience gain after each $m$ direct update. The condition $S^{x} F^{y} \geqslant F^{x} S^{y}$ can be interpreted as follows: compared with $y$, $x$ tends to have a higher trust gain in $A$ after each turn via direct experience. Note that $t^x$ and $t^y$ can be computed exactly by solving a cubic equation or readily approximated by Newton's method. Details are given in the appendix.

A special case is when $n=0$, i.e., agent $x$ only updates trust in $A$ via direct experience, and agent $y$ only updates trust via indirect experience. Theorem~\ref{thm:equilibria} leads to the following corollary with a closed-form equilibrium:
\begin{corollary}
\label{thm:converge_n=0}
When $m >0$ and $n=0$, $x$'s trust $t_{k}^{x,A}$ in $A$ converges to $t^{x} =\frac{s^{x,A} r}{f^{x,A}\overline{r} +s^{x,A} r}$ in probability, i.e., for any $\epsilon  >0$,
\begin{equation*}
\lim _{k\rightarrow \infty } \Pr\left(\left| t_{k}^{x,A} -\frac{s^{x,A} r}{f^{x,A}\overline{r} +s^{x,A} r}\right|  >\epsilon \right) =0.
\end{equation*}
The difference between $t_{k}^{x,A}$ and $t_{k}^{y,A}$ converge to 0 in probability, i.e., for any $\epsilon  >0$,
\begin{equation*}
\lim _{k\rightarrow \infty } \Pr\left(\left| t_{k}^{x,A} -t_{k}^{y,A}\right|  >\epsilon \right) =0.
\end{equation*}
Equivalently, we have $t^{x} =t^{y}$ in Theorem~\ref{thm:equilibria}.
\end{corollary}

Corollay \ref{thm:converge_n=0} implies that under direct-only trust updates, $x$'s trust will stabilize around the following closed-form
$$\frac{s^{x,A} r}{f^{x,A}\overline{r} +s^{x,A} r},$$ which is determined by $x$'s unit experience gains $s^{x,A}$ and $f^{x,A}$ via direct trust, and the robot's reliability $r$; moreover, under indirect-only updates, $y$'s trust will converge to $x$'s trust.

\subsection{Parameter Inference}\label{sec:para_infer}
The proposed model characterizes a human agent's trust in a robot by six parameters. For instance, the parameter of $x$ on robot $A$, which is defined as
\begin{equation}
\label{eq:theta_def}
\vb*{\theta} ^{x,A} =\left( \alpha _{0}^{x,A} ,\beta _{0}^{x,A} ,s^{x,A} ,f^{x,A} ,\hat{s}^{x,A} ,\hat{f}^{x,A}\right) ,
\end{equation}
includes $x$'s prior experiences $\alpha _{0}^{x,A}$ and $\beta _{0}^{x,A}$, the unit direct experience gains $s_{A}^{x}$ and $f_{A}^{x}$, and the unit indirect experience gains $\hat{s}_{A}^{x}$ and $\hat{f}_{A}^{x}$. Denote the indices of $x$'s direct and indirect interactions with $A$ up to time $k$ as $D_{k}$ and $\overline{D}_{k}$. We can compute $\alpha _{k}^{x,A}$ and $\beta _{k}^{x,A}$, according to Eqs.~\eqref{eq:direct_update} and \eqref{eq:indirect_update}, as
\begin{equation}
\label{eq:alpha_k_beta_k}
\begin{aligned}
\alpha _{k}^{x,A} = & \alpha _{0}^{x,A} +s^{x,A}\sum _{j\in D_{k}} p_{j}^{A} +\hat{s}\sum _{j\in \overline{D}_{k}} t_{j}^{x,y}\left[ t_{j}^{y,A} -t_{j-1}^{x,A}\right]^{+}\\
\beta _{k}^{x,A} = & \beta _{0}^{x,A} +f^{x,A}\sum _{j\in D_{k}}\overline{p}_{j}^{A} +\hat{f}\sum _{j\in \overline{D}_{k}} t_{j}^{x,y}\left[ t_{j-1}^{x,A} -t_{j}^{y,A}\right]^{+}
\end{aligned} .
\end{equation}
We compute the optimal parameter $\vb*{\theta} ^{x,A}_{*}$ by maximum likelihood estimation (MLE), i.e., 
\begin{equation*}
\begin{aligned}
\vb*{\theta}^{x,A}_{*} = & \arg\max\log\Pr\left(\text{data}\middle| \vb*{\theta} ^{x,A}\right)\\
= & \arg\max\sum _{k=0}^{K}\log\operatorname{Beta}\left( t_{k}^{x,A}\middle| \alpha _{k}^{x,A} ,\beta _{k}^{x,A}\right)\text{.}
\end{aligned}
\end{equation*}
Specifically, the problem of estimating $x$'s parameter $\vb*{\theta} _{*}^{x,A}$ on robot $A$ is formulated as follows: given $x$'s full trust history in $A$, $\{t_{k}^{x,A}\}_{k=0,1,\dotsc ,K}$, $A$'s performance history during $x$'s direct trust update in $A$, $\{( p_{k}^{A} ,\overline{p}_{k}^{A})\}_{k\in D_{K}}$, $x$'s trust in $y$ during $x$'s indirect trust update in $A$, $\{t_{k}^{x,y}\}_{k\in \overline{D}_{K}}$, and $y$'s trust in $A$ during $x$'s indirect trust update in $A$, $\{t_{k}^{y,A}\}_{k\in \overline{D}_{K}}$, we compute the parameter $\vb*{\theta} _{*}^{x,A}$ that maximizes the log likelihood function 
\begin{equation}
\label{eq:likelihood_def}
H( \vb*{\theta} ^{x,A}) :=\sum _{k=0}^{K}\log\operatorname{Beta}\left( t_{k}^{x,A}\middle| \alpha _{k}^{x,A} ,\beta _{k}^{x,A}\right) ,
\end{equation}
where $\alpha _{k}^{x,A}$ and $\beta _{k}^{x,A}$ are defined in Eq.~\eqref{eq:alpha_k_beta_k}.

We note that $\log\operatorname{Beta}( t_{k}^{x,A} | \alpha _{k}^{x,A} ,\beta _{k}^{x,A})$ is concave in $\vb*{\theta} ^{x,A}$ because it is concave in $( \alpha _{k}^{x,A} ,\beta _{k}^{x,A})$ and $\alpha _{k}^{x,A}$ and $\beta _{k}^{x,A}$ are non-decreasing linear functions of $\vb*{\theta} ^{x,A}$. Consequently, $H( \vb*{\theta} ^{x,A})$ is concave in $\vb*{\theta} ^{x,A}$ since it is a summation of several concave functions. Therefore, we can run the gradient descent method to compute the optimal parameters. 

Now we explicitly give the formulas for the gradient descent method. By expressing the probability density function of Beta random variables in terms of Gamma functions, we can rewrite Eq.~\eqref{eq:likelihood_def} as
\begin{equation*}
\begin{aligned}
 & H( \vb*{\theta}^{x,A} )\\
= & \sum _{k=0}^{K}\left[\log \Gamma \bigl( \alpha _{k}^{x,A} +\beta _{k}^{x,A}\bigr) -\log \Gamma \bigl( \alpha _{k}^{x,A}\bigr) -\log \Gamma \bigl( \beta _{k}^{x,A}\bigr)\right. \\
 & \left. +\bigl( \alpha _{k}^{x,A} -1\bigr)\log t_{k}^{x,A} +\bigl( \beta _{k}^{x,A} -1\bigr)\log\bigl( 1-t_{k}^{x,A}\bigr)\right] ,
\end{aligned}
\end{equation*}
where $\Gamma(\cdot) $ stands for the Gamma function. Define the following variables:
\begin{equation*}
\begin{aligned}
P_{k} & :=\sum _{j\in D_{k}} p_{j}^{A} , & Q_{k} & :=\sum _{j\in \overline{D}_{k}} t_{j}^{x,y}\left[ t_{j}^{y,A} -t_{j-1}^{x,A}\right]^{+},\\
\overline{P}_{k} & :=\sum _{j\in D_{k}}\overline{p}_{j}^{A} , & \overline{Q}_{k} & :=\sum _{j\in \overline{D}_{k}} t_{j}^{x,y}\left[ t_{j-1}^{x,A} -t_{j}^{y,A}\right]^{+}.
\end{aligned}
\end{equation*}
Then \eqref{eq:alpha_k_beta_k} becomes
\begin{equation*}
\begin{aligned}
\alpha _{k}^{x,A} = & \alpha _{0}^{x,A} +s^{x,A} P_{k} +\hat{s}^{x,A} Q_{k}\\
\beta _{k}^{x,A} = & \beta _{0}^{x,A} +f^{x,A}\overline{P}_{k} +\hat{f}^{x,A}\overline{Q}_{k}
\end{aligned} .
\end{equation*}

Calculation shows the gradient can be written as 
\begin{equation}
\label{eq:H_gradient}
\nabla H( \vb*{\theta} ^{x,A}) =\sum _{k=0}^{K} \vb{C}_{k} \vb{v}_{k} ,
\end{equation}
where
\begin{equation}
\label{eq:Ck}
\vb{C}_{k} =\begin{bmatrix}
1 & -1 & 0 & 1 & 0\\
1 & 0 & -1 & 0 & 1\\
P_{k} & -P_{k} & 0 & P_{k} & 0\\
\overline{P}_{k} & 0 & -\overline{P}_{k} & 0 & \overline{P}_{k}\\
Q_{k} & -Q_{k} & 0 & Q_{k} & 0\\
\overline{Q}_{k} & 0 & -\overline{Q}_{k} & 0 & \overline{Q}_{k}
\end{bmatrix}
\end{equation}
and
\begin{equation}
\label{eq:vk}
\vb{v}_{k} =\begin{bmatrix}
\psi \left( \alpha _{k}^{x,A} +\beta _{k}^{x,A}\right)\\
\psi \left( \alpha _{k}^{x,A}\right)\\
\psi \left( \beta _{k}^{x,A}\right)\\
\log t_{k}^{x,A}\\
\log\left( 1-t_{k}^{x,A}\right)
\end{bmatrix} .
\end{equation}
Here $\psi $ is the digamma function. Note that $\vb{C}_{k}$ is constant throughout the gradient descent while $\vb{v}_{k}$ needs to be computed in every iteration.

\subsection{Trust Estimation}\label{sec:trust_estimation}
In real HRI scenarios, querying human trust after every interaction is impractical as it introduces extra workload and reduces collaboration efficiency. Instead, we consider the case when human trust is only queried after some, but not all, of the interactions. In particular, we are interested in referring the model parameter $\vb*{\theta} ^{x,A}$ defined in Eq.~\eqref{eq:theta_def} with missing trust values and estimating these missing values with the TIP model. 

Specifically, the input of the trust estimation problem is the same as the parameter inference problem in Section~\ref{sec:para_infer}, except that $t_{u}^{x,A}$, $t_{u}^{x,y}$, and $t_{u}^{y,A}$ are missing for $u\in U$, where $U$ is the collection of interactions without trust ratings. We assume $0\notin U$, that is, the initial trust ratings, $t_{0}^{x,y}$, $t_{0}^{y,A}$, and $t_{0}^{x,A}$, are known. The optimal parameter is defined as the maximizer of the log-likelihood given the available data: 
\begin{equation*}
H_{U}( \vb*{\theta} ^{x,A}) :=\sum _{k\in \{0,\dotsc ,K\} \backslash U}\log\operatorname{Beta}\left( t_{k}^{x,A}\middle| \alpha _{k}^{x,A} ,\beta _{k}^{x,A}\right) .
\end{equation*}
Eq.~\eqref{eq:alpha_k_beta_k} implies that computing the experiences $\alpha _{k}^{x,A}$ and $\beta _{k}^{x,A}$ relies on the trust ratings $t_{j}^{x,y}$, $t_{j}^{y,A}$, and $t_{j}^{x,A}$. We approximate them by the following recursive relations:

\begin{equation*}
\hat{t}_{j}^{x,y} =t_{j}^{x,y} ,\ \hat{t}_{j}^{y,A} =t_{j}^{y,A} ,\ \text{and} \ \hat{t}_{j}^{x,A} =t_{j}^{x,A},
\end{equation*}
for $j\notin U$;
\begin{equation}
\label{eq:trust_app}
\hat{t}_{j}^{x,y} =t_{j'}^{x,y} ,\ \hat{t}_{j}^{y,A} =t_{j'}^{y,A} ,\ \text{and} \ \hat{t}_{j}^{x,A} =t_{j'}^{x,A} ,
\end{equation}
for $j\in U$, where $j'=\max\{0,1,\dotsc ,j-1\} \backslash U$. In other words, we use the trust rating from the most recent interactions to approximate the missing values. We note that the index $j'$ is well defined in Eq.~\eqref{eq:trust_app} since we assume the initial trust ratings are known. Now, we can compute $\alpha _{k}^{x,A}$ and $\beta _{k}^{x,A}$ by the approximated trust values as follows
\begin{equation}
\label{eq:alpha_beta_missing_trust}
\begin{aligned}
\alpha _{k}^{x,A} = & \alpha _{0}^{x,A} +s^{x,A}\sum _{j\in D_{k}} p_{j}^{A} +\hat{s}\sum _{j\in \overline{D}_{k}}\hat{t}_{j}^{x,y}\left[\hat{t}_{j}^{y,A} -\hat{t}_{j-1}^{x,A}\right]^{+}\\
\beta _{k}^{x,A} = & \beta _{0}^{x,A} +f^{x,A}\sum _{j\in D_{k}}\overline{p}_{j}^{A} +\hat{f}\sum _{j\in \overline{D}_{k}}\hat{t}_{j}^{x,y}\left[\hat{t}_{j-1}^{x,A} -\hat{t}_{j}^{y,A}\right]^{+}
\end{aligned} .
\end{equation}
Similar to maximizing $H$, we can apply the gradient descent method to find the maximizer $\vb*{\theta} _{*}^{x,A}$ of $H_{U}$. The gradient $\nabla H_{U}$ can be computed in the same way as Eq.~\eqref{eq:H_gradient} except that the summation is over $\{0,\dotsc ,K\} \backslash U$ instead of $\{0,\dotsc ,K\}$, i.e., 
\begin{equation*}
\nabla H_{U}( \vb*{\theta} ^{x,A}) =\sum _{k\in \{0,\dotsc ,K\} \backslash U} \vb{C}_{k} \vb{v}_{k} ,
\end{equation*}
where $\vb{C}_{k}$ and $\vb{v}_{k}$ are defined in Eqs.~\eqref{eq:Ck} and \eqref{eq:vk} and computed with the estimated trust values.

By substituting $\vb*{\theta} _{*}^{x,A}$ to Eq.~\eqref{eq:alpha_beta_missing_trust}, we can approximate the experiences and further estimate the missing trust rating $t_{u}^{x,A}$ by the expectation $\mu _{u}^{x,A} =\frac{\alpha _{u}^{x,A}}{\alpha _{u}^{x,A} +\beta _{u}^{x,A}}$ for $u\in U$.


\section{Human Subject Study}\label{sec:humanStudy}

We conducted a human-subject experiment with 30 participants to evaluate the TIP model. The experiment, inspired by~\cite{Yang:2017:EEU:2909824.3020230}, simulated a search and detection task where two human agents work with two smart drones to search for threats at multiple sites.

\subsection{Participants}
A total of ${N=30}$ participants (average age = 25.3 years, SD = 4.3 years, 16 females, 14 males) with normal or corrected-to-normal vision formed 15 teams and participated in the experiment. Each participant received a base payment of \$15 and a bonus of up to \$10  depending on their team performance. 


\subsection{Experimental Task and Design}
\begin{figure}[!t]
  \centering
    \begin{subfigure}{1\columnwidth}
    \captionsetup{width=1\linewidth}
        \centering
        \includegraphics[width=1\columnwidth]{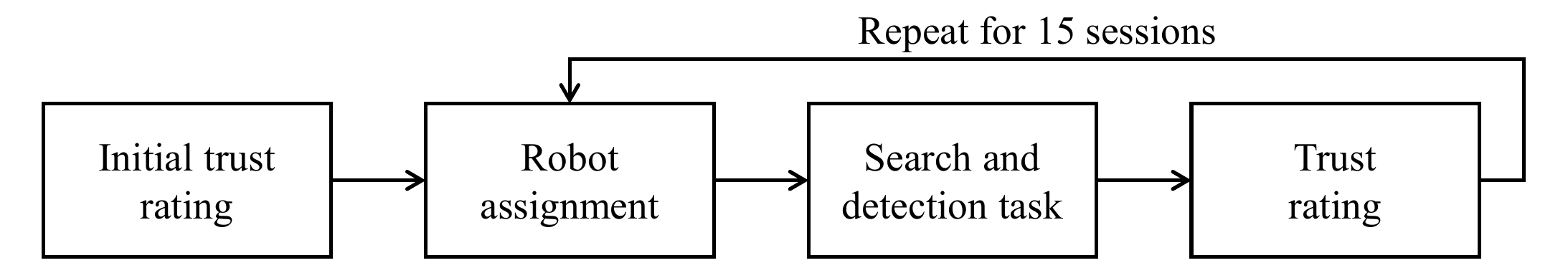}
        \caption{Experiment process.}
        \label{fig:process}
    \end{subfigure}
    \begin{subfigure}{1\columnwidth}
    \captionsetup{width=1\linewidth}
        \centering
        \vspace{5pt}
        \includegraphics[width=1\columnwidth]{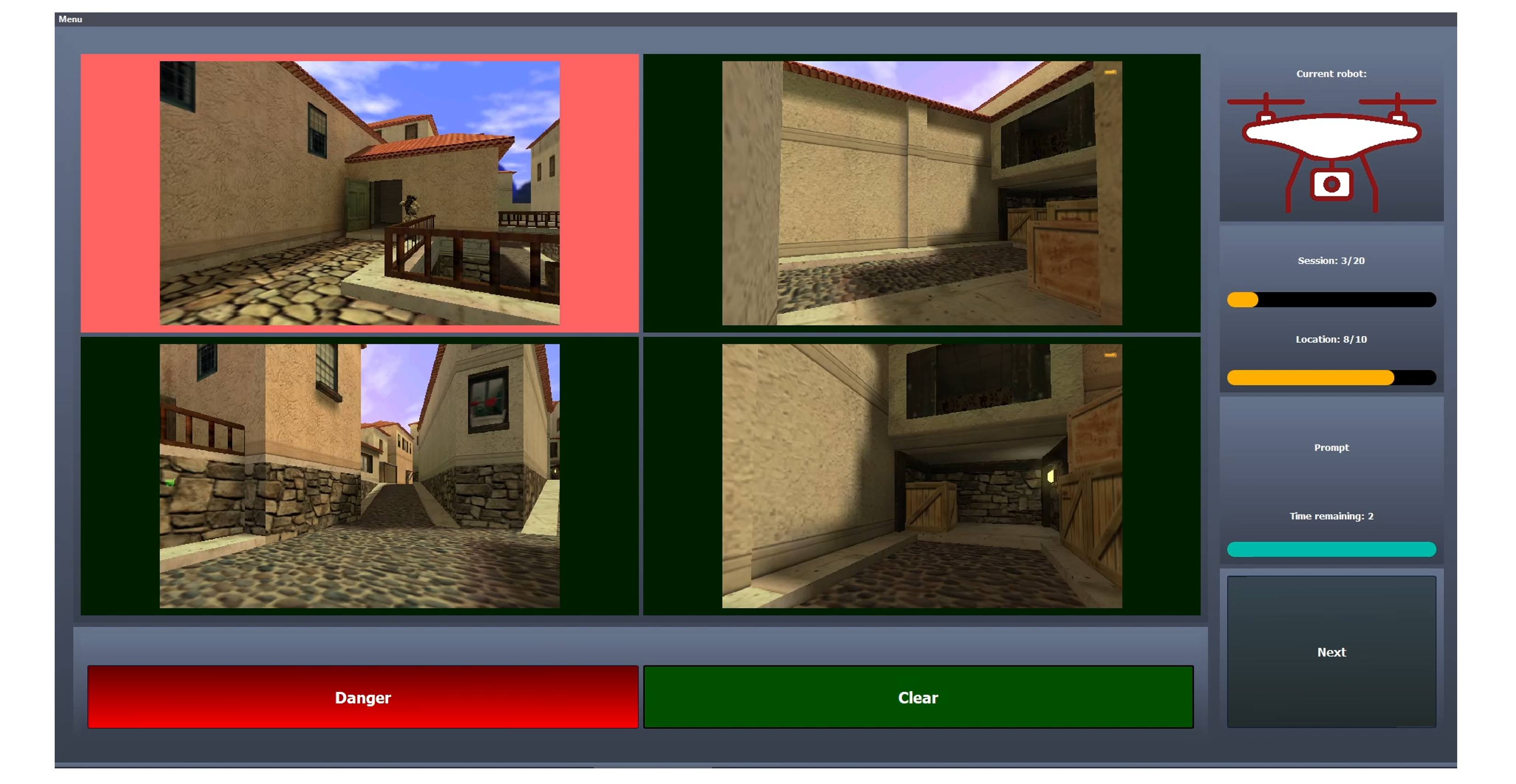}
        \caption{Task interface. The drone will highlight the potential threat in bright red. The participant is asked to click the `Danger' button if a threat is present and to click the `Clear' button otherwise.}
        \label{fig:interface}
    \end{subfigure}
    \begin{subfigure}{1\columnwidth}
    \captionsetup{width=1\linewidth}
        \centering
        \vspace{5pt}
        \includegraphics[width=0.95\columnwidth]{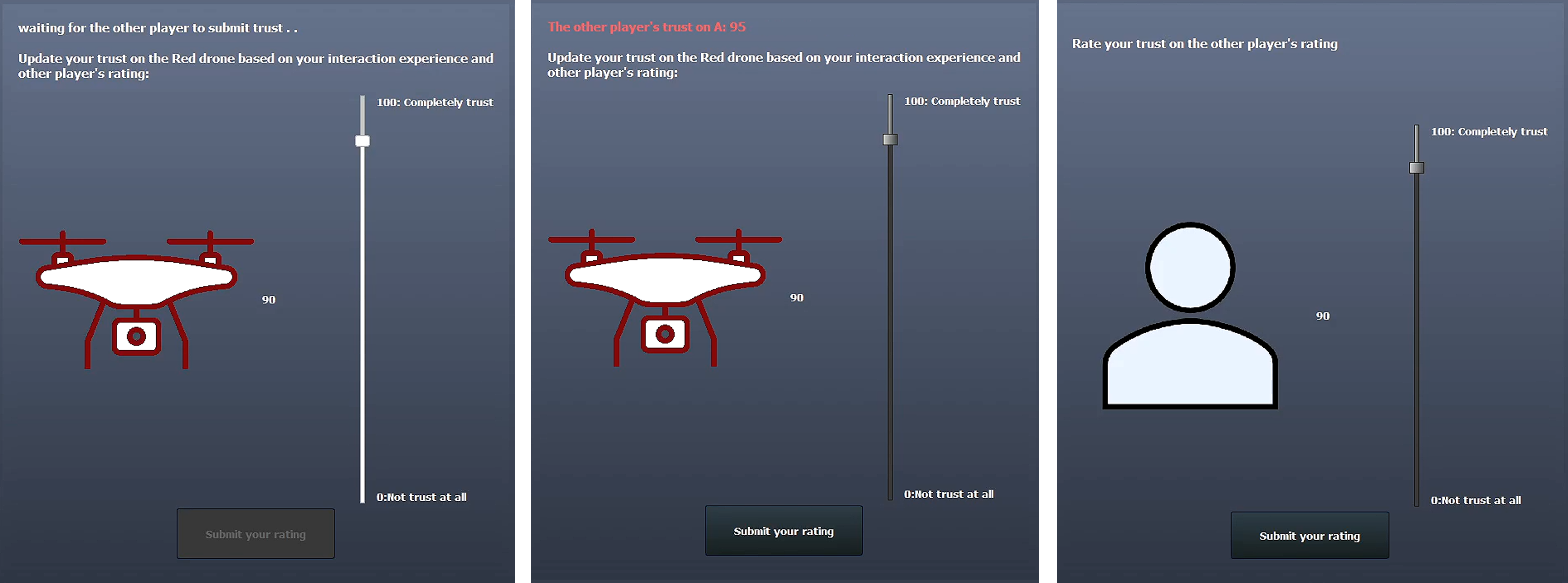}
        \caption{Trust rating interface. Suppose player $x$ was assigned drone $B$ (blue drone). The system first prompted $x$ to rate his or her trust in $B$ based on direct experience and then to update trust in $A$ via indirect experience. The interface of rating $A$ first showed the left figure, where the prompt read ``waiting for the other player to submit trust''. After the other player $y$ rated his or her trust in $A$, the prompt changed the text to ``The other player's trust on $A$: 95'', and asked for player $x$'s rating: ``Update your trust in the red drone based on your interaction experience and the other player's rating'', as shown in the middle figure. Afterward, $x$ would rate trust in the other player $y$, as shown in the right figure.}
        \label{fig:rating_interface}
    \end{subfigure}
    \caption{Experimental process and task interface.}
  \label{fig:different_gaps}
\end{figure}

In the experiment, a pair of participants performed a simulated threat detection task with two assistant drones for $K=15$ sessions on two separate desktop computers. At each session, each participant was assigned one drone and worked on the detection tasks. After the session, they were asked to report their trust in each drone and their trust in their human teammate. 
For clarity, we named the two drones $A$ and $B$ and colored them in red and blue, respectively; and we denoted the participants as $x$ and $y$. A trust rating is denoted as $t^{a,b}_k$, where the superscript $a\in \{x,y\}$ stands for the trustor, the superscript $b\in \{x,y,A,B\}$ stands for the trustee, and the subscript $k$ is the session index. For example, $t^{x,A}_2$ is person $x$'s trust in drone $A$ after the 2nd session. The range of a trust rating is $[0,1]$, where 0 stands for ``(do) not trust at all'' and 1 stands for ``trust completely''. The flow of the experimental task is illustrated in figure~\ref{fig:process}.


\smallskip
\textbf{Initial trust rating.} At the start, each participant gave their initial trust in the two drones based on their prior experience with automation/robots. Additionally, they gave their initial trust in each other. These trust ratings were indexed by 0, e.g., $x$'s initial trust rating in $A$ was denoted as $t^{x,A}_0$.

\begin{figure}[h]
  \centering
  \includegraphics[width=1\linewidth]{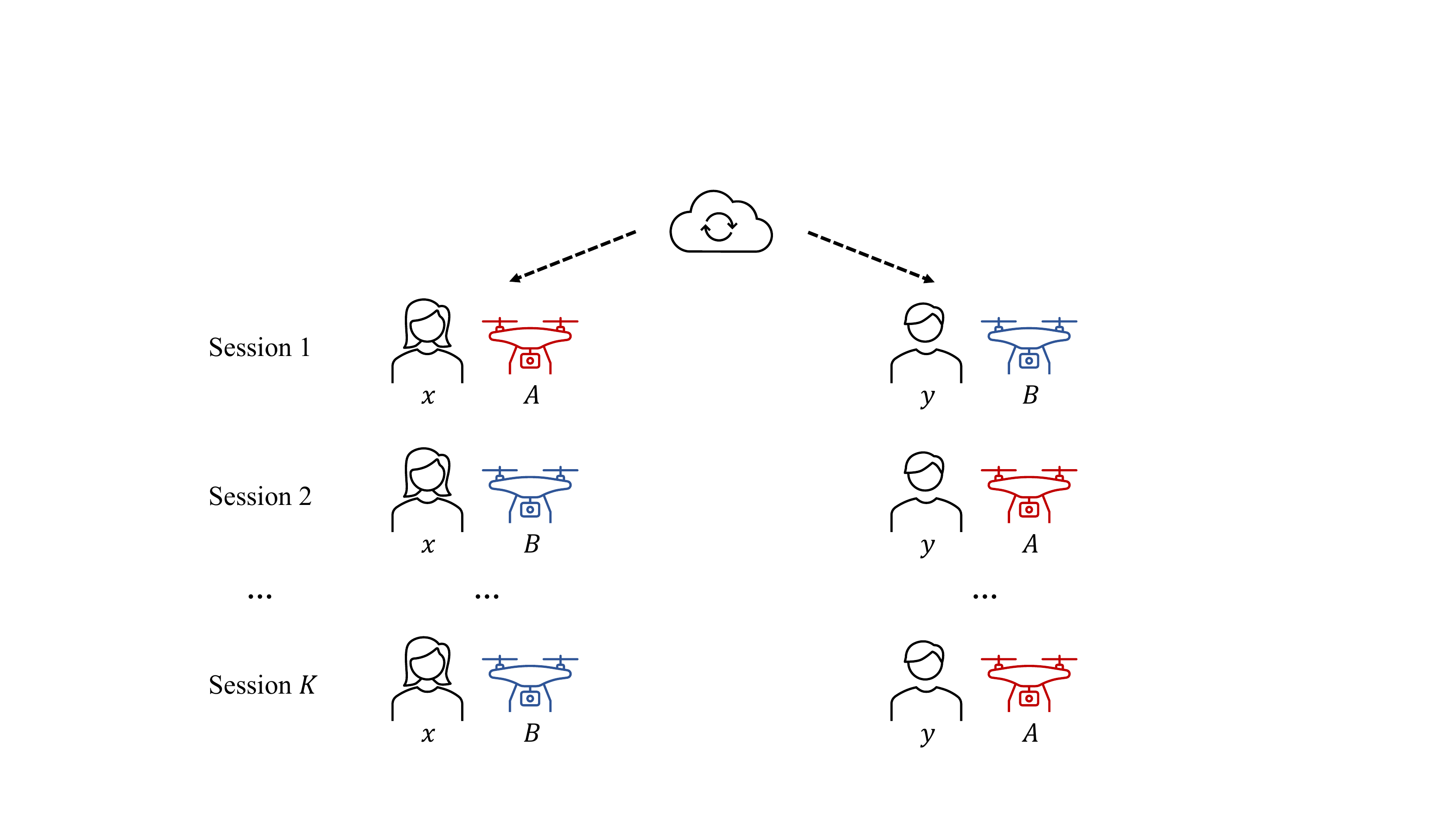}
  \caption{Illustration of drone assignment. Participant $x$ is randomly assigned to work with drone $A$ in session 1, with drone $B$ in session 2, and so on. The assignment is random.}
  \label{fig:drone_assignment}
\end{figure}

\smallskip
\textbf{Robot assignment.} At each session, each participant was randomly assigned one drone as his or her assistant robot, as shown in figure~\ref{fig:drone_assignment}.

\smallskip
\textbf{Search and detection task.} Each session consisted of 10 locations to detect. As shown in figure~\ref{fig:interface}, four views were present at each location. If a threat, which appeared like a combatant, was in any of the views, the participant should click the `Danger' button; otherwise, they should click the `Clear' button. Meanwhile, his or her drone would assist by highlighting a view if the drone detected a threat there. In addition, a 3-second timer was set for each location. If a participant did not click either button before the timer counted down to zero, the testbed
would move to the next location automatically. After all the 10 locations, an end-of-session screen was shown, displaying how many correct choices the participant and the drone had made in the current session. Correct choices mean correctly identifying threats or declaring `Clear' within 3 seconds. 


\smallskip
\textbf{Trust rating.} After each session, each participant reported three trust values.
First, each participant updated his or her trust in the drone s/he just worked with, i.e., through direct experience, based on the drone's detection ability. Next, through a server (see figure~\ref{fig:drone_assignment}), each participant communicated their trust in the drone s/he just worked with to their human teammate. After that, each participant updated his or her trust in the other player's drone (i.e., through indirect experience). Note that only trust ratings were communicated and drones' performances were not. Finally, each participant updated his or her trust in the human teammate based on the teammate’s ability to rate trust in the drones accurately. Hence, after the $k$th session, there would be 6 additional self-reported trust values, $t^{x,A}_k$, $t^{x,B}_k$, $t^{y,A}_k$, $t^{y,B}_k$, $t^{x,y}_k$, and $t^{y,x}_k$. An illustration of the rating interface is shown in figure~\ref{fig:rating_interface}. After participants completed all 15 sessions, the experiment ended.

\subsection{Experimental Procedure}
Prior to the experiment, participants were instructed not to engage in any interaction with each other. Initially, each participant signed a consent form and filled in a demographic survey. To familiarize themselves with the setup, two practice sessions were provided, wherein a practice drone was used to assist the participants. The participants were informed that the practice drone differed from the two drones used in the real experiment. After the experiment started, the assignment of drones was randomized for each pair of participants. 

To motivate participants to provide accurate trust ratings, team performance instead of individual performance was used to determine the bonus, which was calculated as $\$10\times \max\{0,(\Bar{a}-0.7)/0.3\}$, where $\Bar{a}$ was the average detection accuracy of the two participants over all the tasks. Specifically, the participants would receive a bonus if their average detection accuracy exceeded $70\%$. Participants were explicitly informed that truthful and accurate communication of their trust values would assist the other participant in determining the appropriate level of trust in the drones, thereby increasing their detection accuracy and potential bonus.



 \section{Results and Discussion}\label{sec:results_discuccion}
\subsection{Analysis of Trust Convergence within Teams}
We first conduct two types of team-level analysis to demonstrate that leveraging both direct and indirect interaction with a robot leads to faster trust convergence at the team level. We then compare the with- vs. between-team trust deviation and illustrate statistically the existence and benefits of leveraging both direct and indirect experience for trust updating. We denote the set of participants as $P=\{x_1,y_1,\dots,x_{15},y_{15}\}$, where $x_i$ and $y_i$ are two members in the $i$th team.

 \smallskip
\textbf{Within-team trust average over time.} We calculate the within-team trust average for team $i$ on drone $R$ at session $k$ as
\begin{equation*}
\label{eq:team-average}
{t}^{i, R}_k:=\frac{1}{2} ( t_{k}^{x_{i},R} +  t_{k}^{y_{i},R}),
\end{equation*}
where ${R\in \{A,B\}}$ indicates the drone type. The within-team trust average represents a team of players' overall trust in a robot.


\begin{figure}[h]
    \centering
   \includegraphics[width=1\columnwidth]{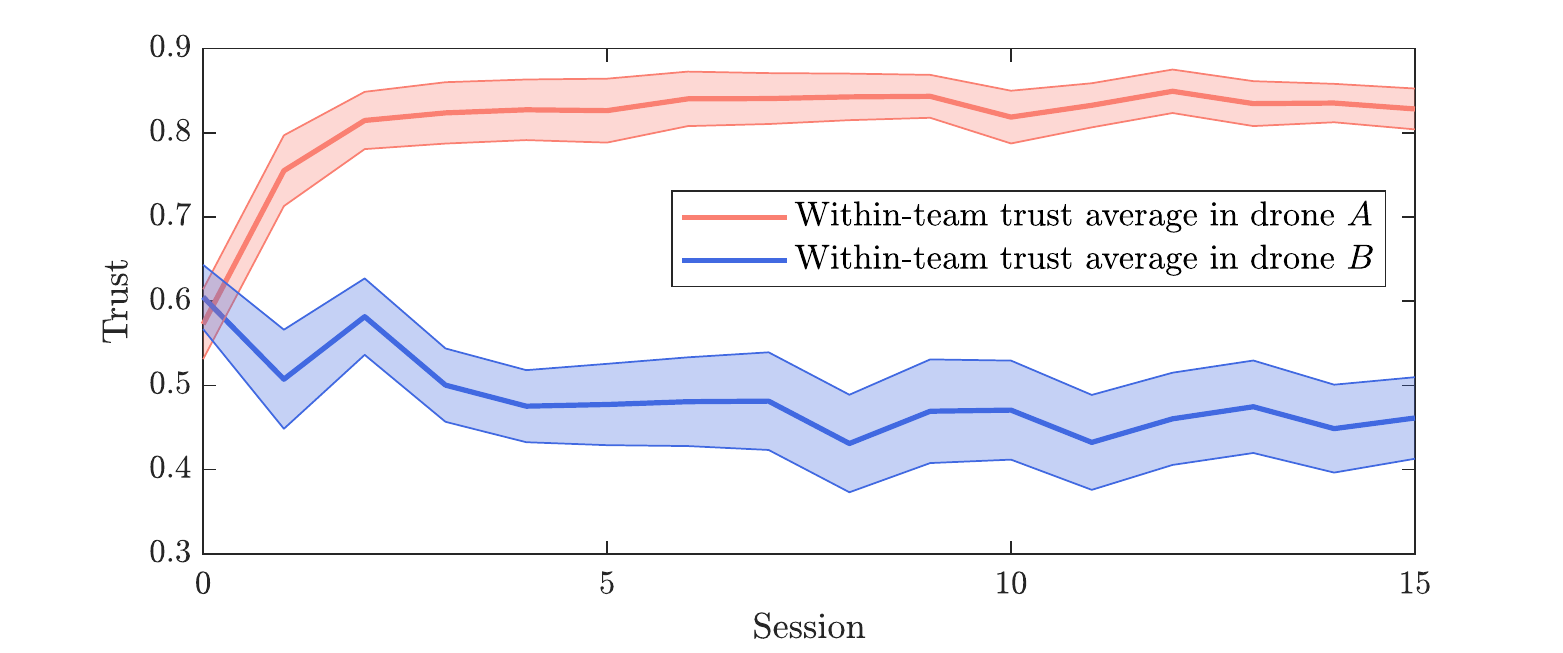}
\caption{Within-team trust \textit{average} in drones $A$ and $B$ over time. Solid lines indicate mean values and the shaded regions indicate the sample standard errors.}
\label{fig:trustbysession}

\end{figure}

Figure~\ref{fig:trustbysession} shows how the within-team average trust changed as the number of interactions increased. The initial and final trusts in drone $A$ ($\frac{1}{15} \sum _{i=1}^{15} {t}^{i, A}_0$ and $\frac{1}{15} \sum _{i=1}^{15} {t}^{i, A}_{15}$) were $0.57 \pm 0.16$ (mean $\pm$ SD) and ${0.83 \pm 0.09}$, respectively. The initial and final trusts in drone $B$ ($\frac{1}{15} \sum _{i=1}^{15} {t}^{i, B}_0$ and $\frac{1}{15} \sum _{i=1}^{15} {t}^{i, B}_{15}$) were ${0.61 \pm 0.15}$ and $0.46 \pm 0.19$, respectively. A two-way repeated measures analysis of variance (ANOVA) showed a significant main effect of drone type (drone $A$ vs. $B$, ${F(1,14)=58.81}$, ${p<.001}$), and a non-significant effect of time (initial vs. final, ${F(1,14)= 3.66}$, ${p=.08}$). There was also a significant interaction effect (${F(1, 14)=73.02}$, ${p<.001}$). Prior to the experiment, the within-team average trust in drone $A$ and that in drone $B$ were similar. As the amount of interaction increased, the within-team average trust in drones $A$ and $B$ tended to reflect the different detection accuracy of drone $A$ and drone $B$, which were set to 90\% and 60\%, respectively. The within-team average trust in drone $A$ gradually increased and that in drone $B$ decreased. At the end of the experiment, the within-team average trust in drone $A$ was significantly larger than that in drone $B$ (${p<0.001}$).

\smallskip
\textbf{Within-team trust deviation over time.} We define the within-team trust deviation of team $i$ on drone $R$ at session $k$ as the difference in trust ratings between the two human players in a team, regardless of whether the trust update is due to direct or indirect interaction, calculated as
\begin{equation*}
\text{dev}^{i, R}_{k,\text{W/N}}:=\lvert t_{k}^{x_{i},R} - t_{k}^{y_{i},R} \lvert, 
\end{equation*}
where $R\in \{A,B\}$ is the drone type and the subscript ``$\text{W/N}$'' stands for ``within.'' In contrast to the within-team trust average, the within-team trust deviation focuses on the differences between the two players in a team. 



\begin{figure}[h]
    \centering
    \includegraphics[width=0.9\columnwidth]{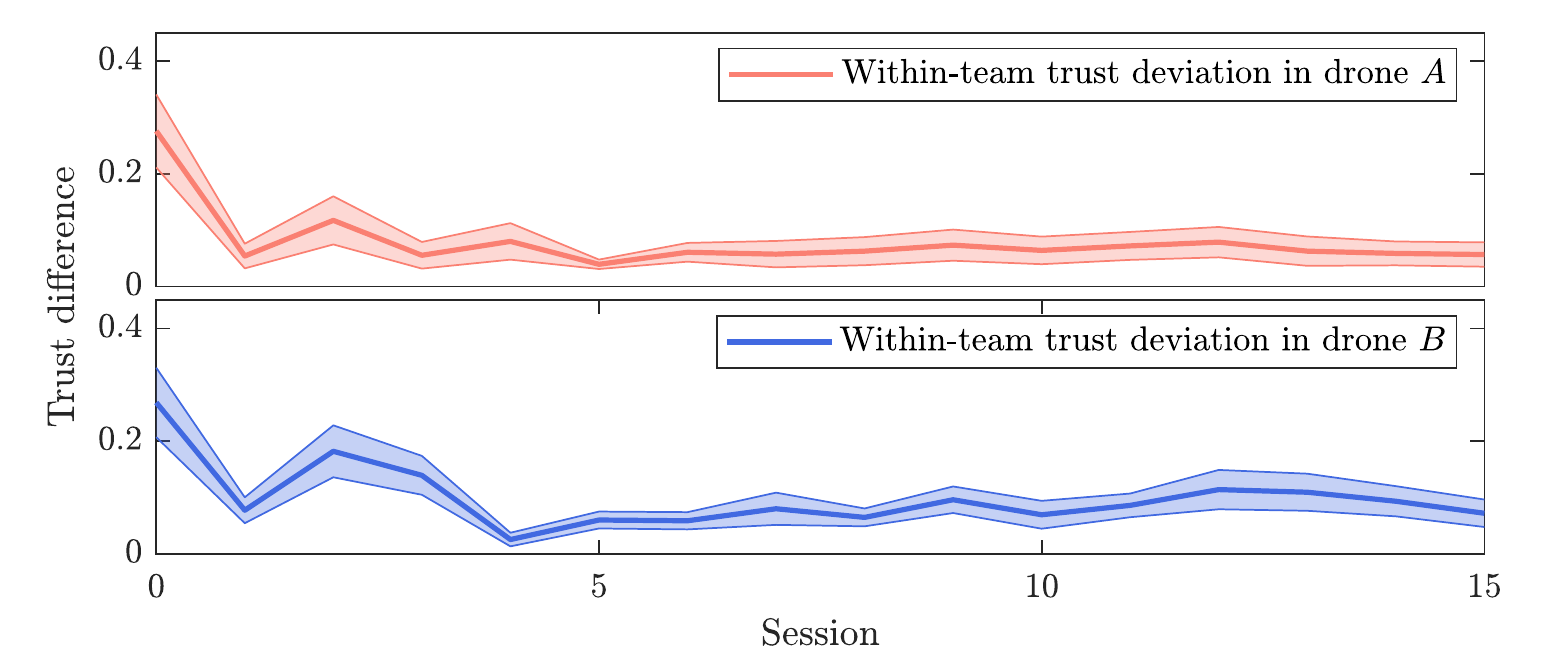}
    \caption{Within-team trust \textit{deviation} in drones $A$ and $B$ over time. Solid lines indicate the mean values and the shaded regions indicate the sample standard errors.}
    \label{fig:trust_deviation_bysession}
\end{figure}

Figure \ref{fig:trust_deviation_bysession} plots the within-team trust deviation in drone $A$ and drone $B$. For both drones, the within-team trust deviation decreased rapidly in the first few sessions and became relatively stable afterward. For drone $A$, the initial and final within-team trust deviations ($\frac{1}{15} \sum_{i=1}^{15} \text{dev}^{i, A}_{0,\text{W/N}}$ and $\frac{1}{15} \sum_{i=1}^{15} \text{dev}^{i, A}_{15,\text{W/N}}$)
were ${0.27 \pm 0.25}$ and 
${0.06\pm0.08}$. For drone $B$, the initial and final trust deviation values ($\frac{1}{15} \sum_{i=1}^{15} \text{dev}^{i, B}_{0,\text{W/N}}$ and $\frac{1}{15} \sum_{i=1}^{15} \text{dev}^{i, B}_{15,\text{W/N}}$) were $0.27 \pm 0.24$ and $0.07 \pm 0.09$. A two-way repeated measures ANOVA revealed a significant main effect of time, that the within-team trust deviation at the end of the experiment was significantly smaller than that prior to the experiment (${F(1,14)=11.51}$, ${p=.004}$). Neither the drone type (${F(1,14)=.06}$, ${p=.82}$) nor the interaction effect (${F(1,14)=.313}$, ${p=.59}$) was significant.

\begin{figure}[h]
    \centering
    \includegraphics[width=0.8\columnwidth]{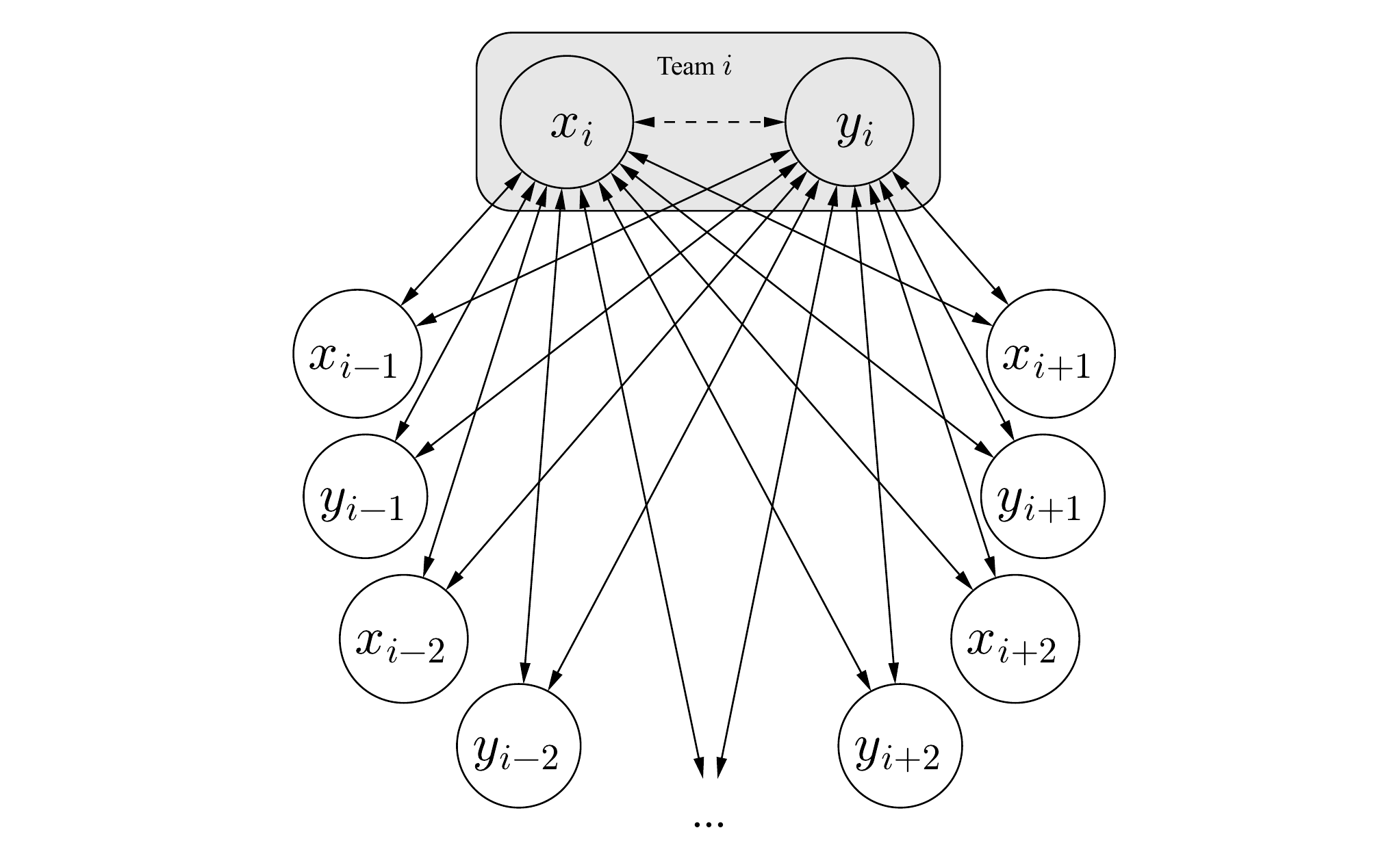}
    \caption{Within-team trust deviation of team $i$ is the trust difference between $x_i$ and $y_i$, indicated by the dashed line in the figure. Between-team trust deviation of team $i$ is the average trust difference between the trust ratings of $x_i$ and $y_i$ and all the other participants in other teams, indicated by the solid lines.}
    \label{fig:cal_between_within}
\end{figure}

\smallskip
\textbf{Within- vs. between-team trust deviation.} To statistically show the existence of trust propagation among team members, we compare the within-team and between-teams trust deviations as human agents gain more interaction experience. If trust propagation between the two players in a team had not occurred (i.e., participants updated their trust in the drones based solely on direct interaction), the within-team and between-team trust deviation would be statistically equal throughout the entire experiment. The between-team trust deviation of the $i$th team on drone $R$ after the $k$th session is defined as
\begin{equation*}
\begin{aligned}
 & \text{dev}_{k,\text{BTW}}^{i,R}\\
:= & {\frac{1}{N-2}\sum _{p\in P\backslash \{x_{i} ,y_{i} \}}\frac{1}{2}\left( |t_{k}^{x_{i} ,R} -t_{k}^{p,R} |+|t_{k}^{y_{i} ,R} -t_{k}^{p,R} |\right)} ,
\end{aligned}
\end{equation*}
where $R\in \{A,B\}$ and $N$ was the total number of participants. Figure \ref{fig:cal_between_within} illustrates the calculation of within- and between-team trust deviations.

\begin{figure}[h]
    \centering
    \includegraphics[width=1\columnwidth]{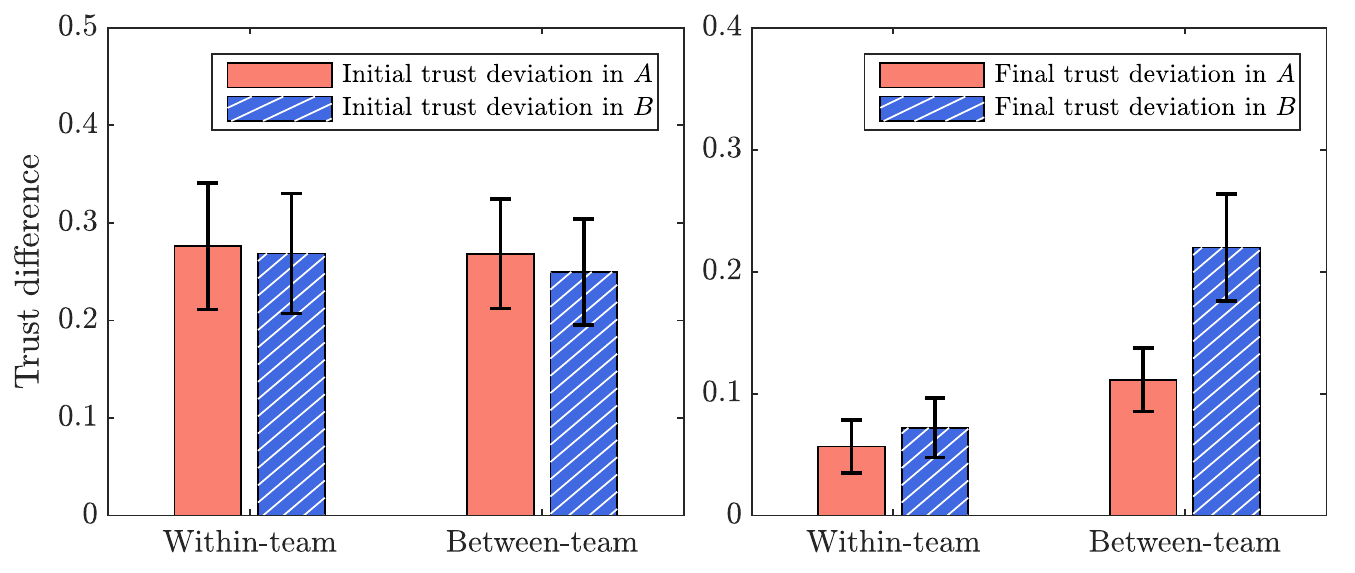} 
    \caption{Within- vs. between-team trust deviations (a) at the beginning and (b) end of the experiment}
    \label{fig:betweenVSwithin}
\end{figure}

Figure~\ref{fig:betweenVSwithin} shows the within- vs. between-team trust deviations at the beginning and end of the experiment. In the beginning, the within- and between-team trust deviations in drone $A$ were $0.28 \pm 0.25$ and $0.27 \pm 0.22$, respectively, and in drone $B$ were $0.27 \pm 0.24$ and $0.25 \pm 0.21$, respectively (figure \ref{fig:betweenVSwithin}(a)). A two-way repeated measures ANOVA showed no significant difference between the within- and between-team trust deviation (${F(1, 14)=.07}$, ${p=.90}$). No difference was found between drone $A$ and drone $B$ (${F(1,14)=2.82}$, ${p=.12}$). The interaction effect was not significant either (${F(1, 14)=0.75}$, ${p=.40}$). 

At the end of the experiment, the within-team and between-team trust deviations in drone $A$ were $0.06 \pm 0.08$ and $0.11 \pm 0.04$, and in drone $B$ were $0.07 \pm 0.09$ and $0.22 \pm 0.08$ (figure \ref{fig:betweenVSwithin}(b)). A two-way repeated measures ANOVA revealed that the within-team trust deviation is significantly smaller than the between-team deviation (${F(1, 14)=71.16}$, ${p<.001}$), and trust deviation in drone $A$ is significantly smaller than drone $B$ (${F(1, 14)=9.81}$, ${p=.007}$). In addition, there was also a significant interaction effect (${F(1, 14)=5.86}$, ${p=.03}$). 

The above results demonstrate the existence, and more importantly, the benefits of trust propagation. As shown in figures \ref{fig:trustbysession} and \ref{fig:trust_deviation_bysession}, the within-team trust average quickly stabilized and the within-team trust deviation rapidly decreased because of trust propagation within a team. Statistically speaking, at the beginning of the experiment, the within-team and between-team trust deviation in both drones were not significantly different (see figure \ref{fig:betweenVSwithin}(a)). At the end of the experiment, the within-team trust deviation was significantly smaller than the between-team trust deviation (see figure \ref{fig:betweenVSwithin}(b)). Had there not been trust propagation between the two players in a team (i.e., participants update their trust in the drones based \textit{only} on the direct interaction), the within-team and between-team trust deviations would remain statistically equal. Therefore, the significant difference at the end of the experiment was attributed to the trust propagation within a team. Being able to fuse one's direct and indirect experience, instead of relying solely on the direct experience, contributes to the quick convergence of trust assessments on a robot, leading to a significantly smaller within-team trust deviation compared to the between-team trust deviation.

\begin{figure*}[h!]
  \centering
  \includegraphics[width=1\linewidth]{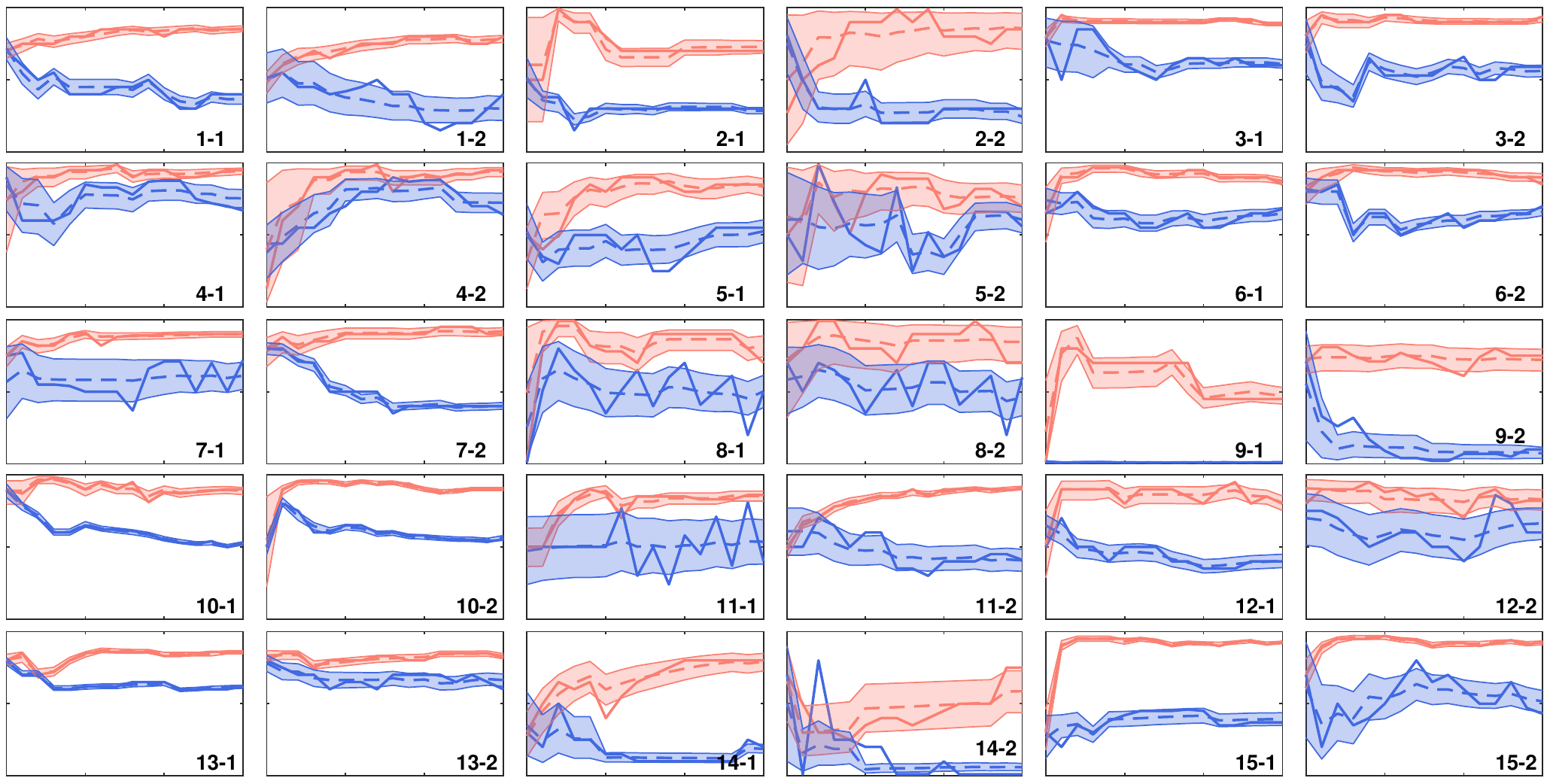}
  \caption{Fitting results. Red curves are for drone $A$ while blue curves are for drone $B$. The solid lines are the participants' trust feedback, while the dashed lines are the expected trust value given by the model. The shaded areas indicate the 90\% probability interval of the Beta distribution at each session. The index $i$-$j$ stands for the $j$th participant in the $i$the group. The horizontal axes represent the session number, ranging from 0 (prior interaction index) to 15. The vertical axes indicate trust levels, ranging from 0 to 1, where 0 represents ``(do) not trust at all'' and 1 indicates ``trust completely''.}
  \label{fig:fitting_all}
\end{figure*}

\subsection{Model Fitting}

\begin{figure}[h!]
    \centering
    \includegraphics[width=1\columnwidth]{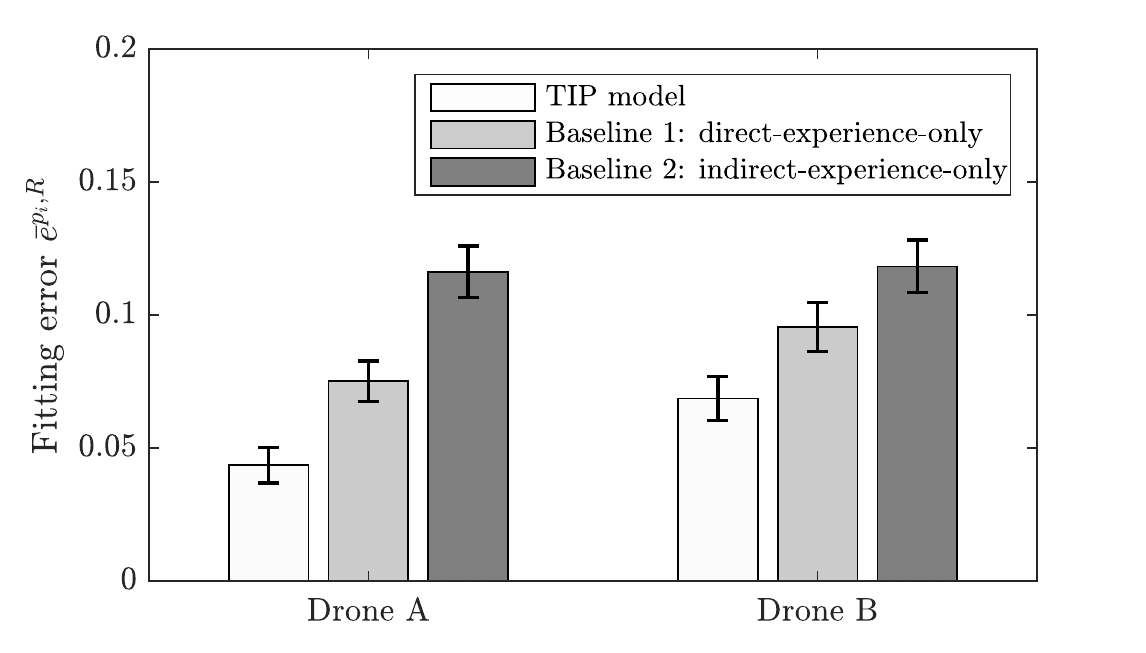} 
    \caption{Fitting error comparison between the TIP model and the two baseline models.}
    \label{fig:fitting_error_bar}
\end{figure}

For clarity, we relabel the participants as $P=\{p_1,p_2,\dots, p_{30}\}$. We utilize the gradient descent method in Section~\ref{sec:para_infer} to compute the optimal parameters ${\vb*{\theta}^{p_i,A}_*}$ and ${\vb*{\theta}^{p_i,B}_*}$ for each participant $p_i$. The fitting results are shown in figure~\ref{fig:fitting_all}. 
We set the performance measurements of drone $A$ at session $k$ as ${p_{k}^{A} =A_{k} /10}$ and ${\overline{p}_{k}^{A} =1-p_{k}^{A}}$, where $A_{k}$ is the number of correct choices drone $A$ made in the $k$th session; and we define $p_{k}^{B}$ and $\overline{p}_{k}^{B}$ similarly. To measure the performance of the model, we define the fitting error at each session for each participant as 
\begin{equation*}
e_{k}^{p_{i} ,R} =\left| \mu _{k}^{p_{i} ,R} -t_{k}^{p_{i} ,R}\right| ,\ R\in \{A,B\},
\end{equation*}
where $t_{k}^{p_{i} ,R}$ is the participant's reported trust while $\mu_{k}^{p_{i} ,R}$ is the expected trust computed according to Eq.~\eqref{eq:trust_expected} with $\alpha _{k}^{p_{i} ,R}$ and $\beta _{k}^{p_{i} ,R}$ generated by Eq.~\eqref{eq:alpha_k_beta_k} based on $\vb*{\theta} _{*}^{p_{i} ,R}$; and, we define the root-mean-square error (RMSE) between the ground truth and the expected trust value as
\begin{equation*}
\text{RMSE}^{R} =\left[\frac{1}{N}\sum _{i=1}^{N}\frac{1}{K+1}\sum _{k=0}^{K}\left( e_{k}^{p_{i} ,R}\right)^{2}\right]^{1/2} ,
\end{equation*}
for ${R\in \{A,B\}}$. The results are ${\text{RMSE}^A=0.057}$ and ${\text{RMSE}^B=0.082}$.


For comparison, we consider two baseline models: one accounting for solely direct experience and another solely indirect experience. The direct-experience-only model corresponds to the TIP model with zero unit gains in indirect experience, i.e., ${\hat{s}^{x,A}=\hat{f}^{x,A}=0}$; while the indirect-experience-only model corresponds to ${{s}^{a,b}={f}^{a,b}=0}$. We recompute the parameters for the baseline models, and the RMSE errors are ${\text{RMSE}^A_{\text{direct}}=0.085}$, ${\text{RMSE}^B_{\text{direct}}=0.107}$, ${\text{RMSE}^A_{\text{indirect}}=0.128}$, and ${\text{RMSE}^B_{\text{indirect}}=0.130}$. In addition, we compare each participant's fitting error ${\bar{e}^{p_{i} ,R}:={1}/({K+1})\sum_{k=0}^K e_{k}^{p_{i} ,R}}$ of the TIP model ($A$: $0.044 \pm 0.037$; $B$: $0.069 \pm 0.045$), direct-experience-only model ($A$: $0.075 \pm 0.041$; $B$: $0.095 \pm 0.051$), and indirect-experience-only model ($A$: $0.116 \pm 0.053$; $B$: $0.118 \pm 0.054$) using a paired-sample t-test. Results reveal that the fitting error of the TIP model is significantly smaller than the direct-experience-only model, with $t(29)=-6.18, p<.001$ for drone $A$, and ${t(29)=-7.31}$, ${p<.001}$ for drone $B$, and significantly smaller than the indirect-experience-only model, with $t(29)=-9.28, p<.001$ for drone $A$, and ${t(29)=-10.06}$, ${p<.001}$ for drone $B$. Furthermore, the fitting error of the direct-only model is significantly smaller than the indirect-experience-only model, with $t(29)=-4.73, p<.001$ for drone $A$, and ${t(29)=-3.73}$, ${p<.001}$ for drone $B$. A bar plot is shown in figure~\ref{fig:fitting_error_bar}. This comparison indicates that a human agent mainly relies on direct experience to update his or her trust, while indirect experience also plays a vital role in trust dynamics.

\subsection{Trust Estimation}
To measure the estimation accuracy of the proposed model, we remove some trust ratings in the data and compute the RMSE of the estimated trust values. Specifically, for each participant $p_{i}$, we set ${U_{\hat{K}} =\{{K-\hat{K} +1},\dotsc ,K\}}$ to remove the last $\hat{K}$ trust ratings, where $U_{\hat{K}}$ is the index set of sessions without trust ratings as defined in Section~\ref{sec:trust_estimation}, and estimate the missing trust values by $\mu _{u}^{p_{i} ,A}$ and $\mu _{u}^{p_{i} ,B}$ for $u\in U_{\hat{K}}$. The root-mean-square errors are defined as
\begin{equation*}
\text{RMSE}_{\hat{K}}^{R} =\left[\frac{1}{N}\sum _{i=1}^{N}\frac{1}{\hat{K}}\sum _{u=K-\hat{K} +1}^{K}\left( e_{u}^{p_{i} ,R}\right)^{2}\right]^{\frac{1}{2}} ,
\end{equation*}
for $R\in \{A,B\}$. 

\begin{figure}[h]
  \centering
  \includegraphics[width=1\linewidth]{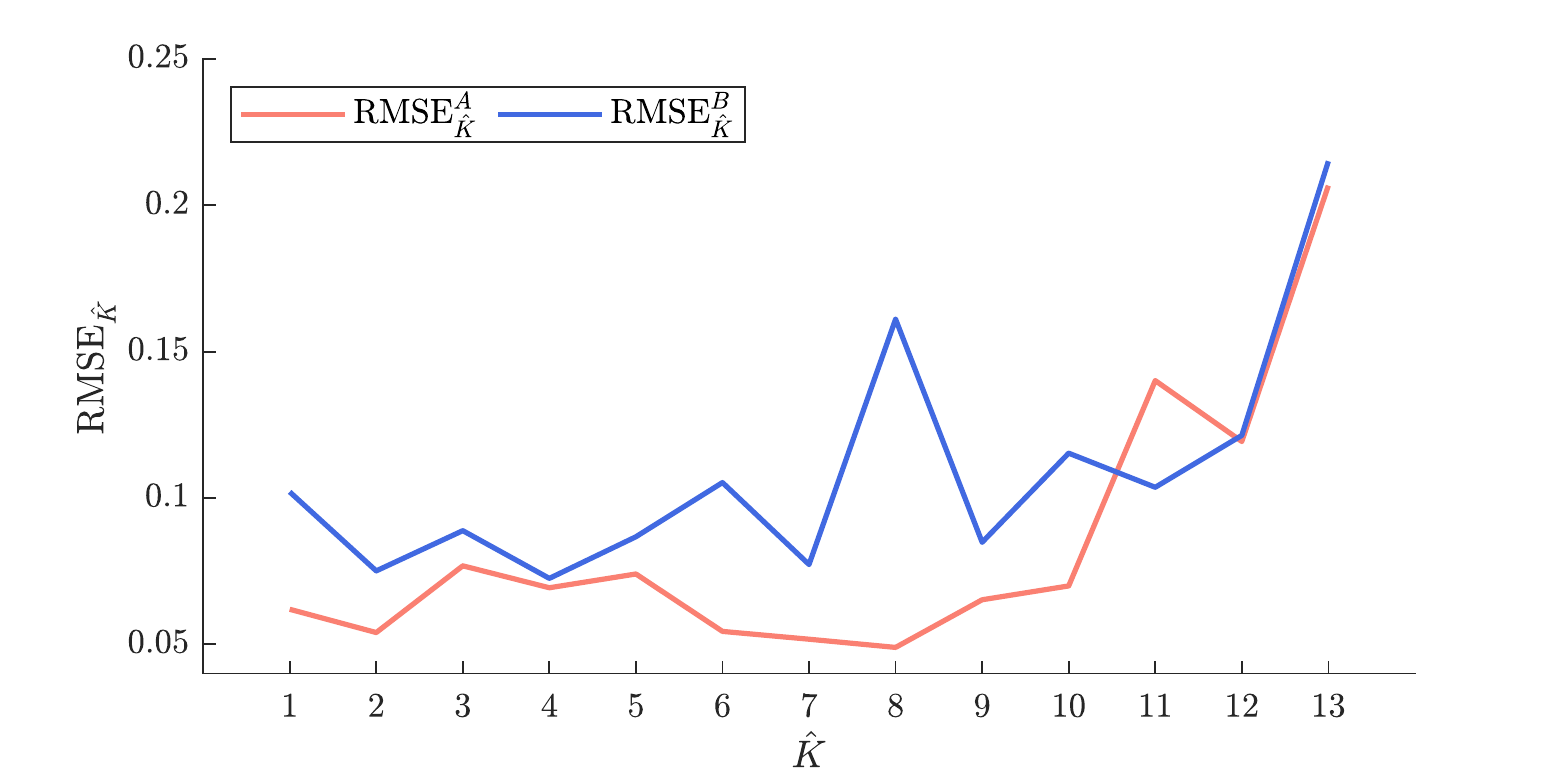}
  \caption{RMSE's under different $\hat{K}$. $\hat{K}$ is the number of interactions without trust feedback. Equivalently, the TIP model uses the trust ratings from the first ${16-\hat{K}}$ sessions to learn the model parameters.  
  }
  \label{fig:RMSE_missing_trust}
\end{figure}

Figure~\ref{fig:RMSE_missing_trust} shows the RMSE's under different $\hat{K}$.
When ${\hat{K}\leq 7}$, the TIP model can successfully estimate the trust values in the late sessions with a small RMSE ($< 0.1$) by learning from previous data. In particular, ${\text{RMSE}_{\hat{K} =7}^{A} =0.052}$ and ${\text{RMSE}_{\hat{K} =7}^{B} =0.077}$, which implies that, with the first 9 sessions' trust ratings available, the RMSE's of the estimation for the last 7 sessions are under 0.08 for both drones. The result also illustrates that $\text{RMSE}_{\hat{K}}^{A}$ is smaller than $\text{RMSE}_{\hat{K}}^{B}$ in general. This could be explained by the performance difference between the two drones. Indeed, because the number of correct choices each drone could make follows binomial distributions ($\operatorname{Bin}(10, 0.9)$ for $A$ and $\operatorname{Bin}(10, 0.6)$ for $B$), the variance of their performance are 0.09 and 0.24 respectively. The greater variance of drone $B$ may cause a human subject to acquire more information to stabilize his or her trust and thus leads to higher uncertainty in trust feedback values, which makes it difficult for the model to learn trust dynamics in a short time.

\section{Conclusion} \label{sec:conclusion}
In the study, we proposed the TIP model that accounts for both the direct and indirect experiences a human agent may have with a robot in multi-human multi-robot teams. To the best of our knowledge, it is the first mathematical framework for computational trust modeling in multi-human multi-robot teams. In addition, we prove theoretically that trust converges after repeated direct and indirect interactions under our TIP framework. Using a human-subject experiment, we showed that being able to fuse one's direct and indirect experiences, instead of relying solely on the direct experience, contributes to the quick convergence of trust in a robot. In addition, we showed that the TIP models significantly outperformed the baseline direct-experience-only model in capturing the trust dynamics in multi-human multi-robot teams. The TIP model can be applied to various human-robot teaming contexts including team of teams~\cite{McChrystal} and multi-echelon networks~\cite{NAP26355}.  In particular, the TIP model can update a human agent's trust in a robot whenever a direct or indirect experience is available and thus can be applied for trust estimation in a network consisting of multiple humans and robots.


Our results should be viewed in light of several limitations. First, we assume that the two human players within a team were cooperative and willing to share their trust in a robot truthfully. In a non-cooperative context where a human player is motivated to cheat, a quick convergence of trust assessment is less likely to occur. Further research is needed to examine the non-cooperative scenario. Second, we used a one-dimensional trust scale in the experiment. Even though this scale has been used in prior literature~\cite{manzey2012human, Yang:2017:EEU:2909824.3020230, Bhat_RAL_2022}, it may not capture the different underlying dimensions of trust. Third, we take an ability/performance-centric view of trust and assume a human agent's trust in a robot is primarily driven by the ability or performance of the robot. Based on research in organizational management,  trust can be influenced by three elements, namely ability, integrity, and benevolence~\cite{mayer1995integrative}. Future research should investigate ways to integrate the benevolence and integrity elements into trust modeling, in particular, for HRI contexts that involve a strong emotional component, for example, educational or home-care robots. Moreover, conducting further ablation studies is essential to comprehensively understand the impact of various factors on the dynamics of trust. For instance, varying the number of sessions versus drone performances would provide insights into the rate at which trust converges.

\section*{Acknowledgments}
This work is supported by the Air Force Office of Scientific Research under Grant No. FA9550-23-1-0044.

\bibliographystyle{IEEEtran}
\bibliography{reference}

\newpage

\section*{Appendix}

\subsection{Proof of Theorem \ref{thm:both_converge}}
\noindent\textbf{Case 1}: $n=0$. 

First, we show that $t_{k}^{x,A}$ converges i.p.. When $n=0$, $x$ gains experience towards $A$ through only direct interaction, and thus, by Eq.~\eqref{eq:direct_update},
\begin{equation}
\label{eq:direct_only_limit}
\begin{aligned}
 & \lim _{k\rightarrow \infty } \mu _{k}^{x,A} =\lim _{k\rightarrow \infty }\frac{\alpha _{k}^{x,A}}{\alpha _{k}^{x,A} +\beta _{k}^{x,A}}\\
= & \lim _{k\rightarrow \infty }\frac{\alpha _{0}^{x,A} +ks^{x,A} r}{\alpha _{0}^{x,A} +\beta _{0}^{x,A} +kf^{x,A}\overline{r} +ks^{x,A} r}\\
= & \frac{s^{x,A} r}{f^{x,A}\overline{r} +s^{x,A} r} .
\end{aligned}
\end{equation}
For any $\epsilon  >0$, by the Markov inequality, 
\begin{equation}
\label{eq:markov}
\begin{aligned}
 & \lim _{k\rightarrow \infty }\Pr(| t_{k}^{x,A} -\mu _{k}^{x,A}| < \epsilon )\\
\leqslant  & \lim _{k\rightarrow \infty }\frac{1}{\epsilon ^{2}}\mathbb{E}[( t_{k}^{x,A} -\mu _{k}^{x,A})^{2}] =0,
\end{aligned}
\end{equation}
where the last equality is true because $\lim _{k\rightarrow \infty }\operatorname{var}( t_{k}^{x,A}) =0$ as $\lim _{k\rightarrow \infty }( \alpha _{k}^{x,A} +\beta _{k}^{x,A}) =\infty $. Let $t^{x} =\frac{s^{x,A} r}{f^{x,A}\overline{r} +s^{x,A} r}$. Eqs. \eqref{eq:direct_only_limit} and \eqref{eq:markov} yield
\begin{equation*}
\lim _{k\rightarrow \infty }\Pr(| t_{k}^{x,A} -t^{x}| < \epsilon ) =1.
\end{equation*}

Second, we show $| t_{k-1}^{y,A} -t_{k}^{x,A}| $ converge i.p. to zero. Suppose this is not true. Then, there exist $\epsilon ,\ \delta  >0$ such that there are infinite many $k$'s such that $\Pr\left(| t_{k-1}^{y,A} -t_{k}^{x,A}|  >\epsilon \right)  >\delta $. The indirect updating rule Eq.~\eqref{eq:indirect_update} implies $\alpha _{k}^{y,A} +\beta _{k}^{y,A}\xrightarrow{\text{i.p.}} \infty $. Consequently, similar to Eq.~\eqref{eq:markov}, we obtain $| t_{k}^{y,A} -\mu _{k}^{y,A}| \xrightarrow{\text{i.p.}} 0$. We consider the following equation:
\begin{equation}
\label{eq:mu_limit_1}
\begin{aligned}
 & \left| \mu _{k+m-1}^{y,A} -t_{k+m}^{x,A}\right| \\
= & \left| \mu _{k-1}^{y,A} -t_{k}^{x,A}\right| \prod _{j=1}^{m}\frac{\left| \mu _{k+j-1}^{y,A} -t_{k+j}^{x,A}\right| }{\left| \mu _{k+j-2}^{y,A} -t_{k+j-1}^{x,A}\right| }.
\end{aligned} 
\end{equation}
As we have shown $t_{k}^{x,A}\xrightarrow{\text{i.p.}} t^{x}$ when $k\rightarrow \infty $, by the continuous mapping theorem, when $k\rightarrow \infty $, 
\begin{equation}
\label{eq:mu_limit_2}
\prod _{j=1}^{m}\frac{\left| \mu _{k+j-1}^{y,A} -t_{k+j}^{x,A}\right| }{\left| \mu _{k+j-2}^{y,A} -t_{k+j-1}^{x,A}\right| }\xrightarrow{\text{i.p.}}\prod _{j=1}^{m}\frac{\left| \mu _{k+j-1}^{y,A} -t^{x}\right| }{\left| \mu _{k+j-2}^{y,A} -t^{x}\right| } .
\end{equation}
An examination of Eqs.~\eqref{eq:trust_expected} and \eqref{eq:indirect_update}  shows
\begin{equation}
\label{eq:mu_limit_3}
\prod _{j=1}^{m}\frac{\left| \mu _{k+j-1}^{y,A} -t^{x}\right| }{\left| \mu _{k+j-2}^{y,A} -t^{x}\right| }\xrightarrow{\text{i.p.}} 0
\end{equation}
when $m\rightarrow \infty $. Eqs. \eqref{eq:mu_limit_1}, \eqref{eq:mu_limit_2}, and \eqref{eq:mu_limit_3} together yield $| \mu _{k+m-1}^{y,A} -t_{k+m}^{x,A}| \xrightarrow{\text{i.p.}} 0$ when both $k$ and $m$ tend to infinity. Thus, $\mu _{k}^{y,A}\xrightarrow{\text{i.p.}} t^{x}$. Because we also showed $\left| t_{k}^{y,A} -\mu _{k}^{y,A}\right| \xrightarrow{\text{i.p.}} 0$, $\left| t_{k}^{y,A} -t^{x}\right| \xrightarrow{\text{i.p.}} 0$. This contradicts our assumption. Therefore, $\left| t_{k-1}^{y,A} -t_{k}^{x,A}\right| $ converge i.p. to zero. Particularly, since $t_{k}^{x,A}\xrightarrow{\text{i.p.}} t^{x}$, $t_{k}^{y,A}\xrightarrow{\text{i.p.}} t^{x}$.

Therefore, both $t_{k}^{x,A}$ and $t_{k}^{y,A}$ converge to $t^{x}$ i.p..

\vspace{2mm}
\noindent\textbf{Case 2}: $n >0$.

First, when $k\rightarrow \infty $, $\alpha _{k}^{x,A}$, $\beta _{k}^{x,A}$, $\alpha _{k}^{y,A}$, and $\beta _{k}^{y,A}$ all go to infinity as both $x$ and $y$ will have an infinite number of direct interactions with $A$. As a result, we have
\begin{equation}
\label{eq:t_deviation_go_to_zero}
\begin{aligned}
\left| t_{k}^{x,A} -\mu _{k}^{x,A}\right|  & \xrightarrow{\text{i.p.}} 0 & \text{and}\\
\left| t_{k}^{y,A} -\mu _{k}^{y,A}\right|  & \xrightarrow{\text{i.p.}} 0. & 
\end{aligned}
\end{equation}

Second, let $\Delta t_{k} :=t_{k}^{x,A} -t_{k}^{y,A}$. With the similar technique used in Eqs. \eqref{eq:mu_limit_1}, \eqref{eq:mu_limit_2}, and \eqref{eq:mu_limit_3}, it can be shown that
\begin{equation}
\label{eq:t_diff_converge}
\Delta t_{k}\xrightarrow{\text{i.p.}} \Delta t,
\end{equation}
where $\Delta t$ is some constant.

Finally, we show $t_{k}^{x,A}$ and $t_{k}^{y,A}$ converge to constants i.p.. Since $\left| \mu _{k}^{x,A} -\mu _{k-1}^{x,A}\right| \xrightarrow{\text{i.p.}} 0$, by Eqs. \eqref{eq:t_deviation_go_to_zero} and \eqref{eq:t_diff_converge}, we obtain
\begin{equation*}
\left| t_{k}^{y,A} -\mu _{k-1}^{x,A}\right| \xrightarrow{\text{i.p.}} \Delta t.
\end{equation*}
By Eqs. \eqref{eq:direct_update} and \eqref{eq:indirect_update}, both indirect and direct experience gains of $y$ converge to some constants i.p. Therefore, the ratio $\frac{\ \alpha _{k}^{y,A}}{\alpha _{k}^{y,A} +\beta _{k}^{y,A}}$ converge to some constants i.p. respectively, i.e., there exist some $t^{y}$ such that $t_{k}^{y,A}\xrightarrow{\text{i.p.}} t^{y}$. Similarly, there exists some $t^{x}$ such that $t_{k}^{x,A}\xrightarrow{\text{i.p.}} t^{x}$.
\qed 

\subsection{Proof of Theorem \ref{thm:equilibria}}
When $n=0$, Eq.~\eqref{eq:equilibria_case1} yields $t^{x} =t^{y} =\frac{s^{x,A} r}{f^{x,A}\overline{r} +s^{x,A} r}$, which agrees with the proof of theorem \ref{thm:both_converge}.

Now we consider the case $n >0$. Let $t^{x}$ and $t^{y}$ be the equilibrium in the statement. If $S^{x} /F^{x} -S^{y} /F^{y} \geqslant 0$, it can be shown that
\begin{equation*}
\lim\limits _{k\rightarrow \infty }\Pr\left( t_{k}^{x,A} -t_{k}^{y,A} \geqslant 0\right) =1.
\end{equation*}
As a result, we have
\begin{equation}
\label{eq:trust_delta}
\left[ t_{k-1}^{y,A} -t_{k}^{x,A}\right]^{+}\xrightarrow{\text{i.p.}} 0\ \text{and} \ \left[ t_{k}^{x,A} -t_{k-1}^{y,A}\right]^{+}\xrightarrow{\text{i.p.}} t^{x} -t^{y} .
\end{equation}
When $k\rightarrow \infty $, by Eqs. \eqref{eq:direct_update}, \eqref{eq:indirect_update}, and \eqref{eq:trust_delta}, we have 
\begin{equation*}
\begin{aligned}
\alpha _{kl+l}^{x,A} -\alpha _{kl}^{x,A} & \xrightarrow{\text{i.p.}} S_{A}^{x}\\
\beta _{kl+l}^{x,A} -\beta _{kl}^{x,A} & \xrightarrow{\text{i.p.}}\hat{F}_{A}^{x}\left( t^{x} -t^{y}\right) +F_{A}^{x}
\end{aligned} ,
\end{equation*}
where $l=m+n$. This implies the trust gains are constant every other $l$ interactions. By Eq. \eqref{eq:trust_def},

\begin{equation*}
t^{x} =\lim _{k\rightarrow \infty }\frac{\ \alpha _{k}^{x,A}}{\alpha _{k}^{x,A} +\beta _{k}^{x,A}} =\frac{S_{A}^{x}}{S_{A}^{x} +\hat{F}_{A}^{x}\left( t^{x} -t^{y}\right) +F_{A}^{x}} ,
\end{equation*}
which proves the first equation in Eq. \eqref{eq:equilibria_case1}. The second equation in Eq. \eqref{eq:equilibria_case1} can be proved similarly.

The proof for the case when $S_{A}^{x} /F_{A}^{x} -S_{A}^{y} /F_{A}^{y} < 0$ is similar.
\qed 

\subsection{Computing $t^x_A$ and $t^y_A$}
We consider the case of Eq. \eqref{eq:equilibria_case1}. When $n=0$, the solution is given by Eq.~\eqref{eq:direct_only_limit}. Assume $n\neq 0$. Solving Eq. \eqref{eq:equilibria_case1} directly  results in two cubic equations of $t^x$ and $t^y$ respectively. An exact solution can be derived from these equations on $[0,1]^2$ but the process can be tedious. A more practical method is to exploit the Newton's method to approximate $t^x$ and $t^y$. For example, letting $z=1-y$, Eqs. in \eqref{eq:equilibria_case1} give
\begin{equation*}
\begin{aligned}
\hat{F}^{x} x^{2} +\hat{F}^{x} zx+\left( F^{x} +S^{x} -\hat{F}^{x}\right) x-S^{x} = & 0, & \text{and}\\
\hat{S}^{y} z^{2} +\hat{S}^{y} zx+\left( S^{y} +F^{y} -\hat{S}^{y}\right) z-F^{y} = & 0. & 
\end{aligned}
\end{equation*}
We can define $f_1(x,z)$ and $f_2(x,z)$ to be the left-hand sides of above equations and define $\mathbf{f} =( f_{1} ,f_{2})$. We have

\vspace{1mm}
\resizebox{0.95\hsize}{!}{%
    \begin{math}
\begin{aligned}
& \mathbf{f} '( x,z)\\
= & \begin{pmatrix}
2\hat{F}^{x} x+\hat{F}^{x} z+F^{x} +S^{x} -\hat{F}^{x} & \hat{F}^{x} x\\
\hat{S}^{y} z & 2\hat{S}^{y} z+\hat{S}^{y} x+S^{y} +F^{y} -\hat{S}^{y}
\end{pmatrix}.
\end{aligned}
    \end{math}%
}
\vspace{2mm}

We solve
\begin{equation*}
    ( x_{k+1} ,z_{k+1})^{T} =( x_{k} ,z_{k})^{T} -(\mathbf{f'})^{-1}\mathbf{f}( x_{k} ,z_{k})
\end{equation*}
iteratively for $x$ and $z$ and obtain $t^x=x$ and $t^y =1-z$ as the equilibrium.


\end{document}